\documentclass[preprint,12pt]{elsarticle}
\usepackage{amssymb}
\usepackage{placeins}
\usepackage{rotating}
\usepackage{colortbl}
\usepackage{lscape}
\usepackage{longtable}
\usepackage{adjustbox}
\usepackage{pdflscape}
\newcolumntype{L}{>{\centering\arraybackslash}m{6cm}}
\newcolumntype{M}{>{\centering\arraybackslash}m{2.5cm}l}

\biboptions{sort&compress}

\journal{Journal of Biomedical Informatics}

\begin{document}

\begin{frontmatter}

\title{Ensemble Machine Learning Model Trained on a New Synthesized Dataset Generalizes Well for Stress Prediction Using Wearable Devices}

\author[inst1]{Gideon Vos}

\affiliation[inst1]{organization={College of Science and Engineering, James Cook University},
            addressline={James Cook Dr}, 
            city={Townsville},
            postcode={4811}, 
            state={QLD},
            country={Australia}}

\author[inst1]{Kelly Trinh}
\author[inst2]{Zoltan Sarnyai}
\author[inst1]{Mostafa Rahimi Azghadi}

\affiliation[inst2]{organization={College of Public Health, Medical, and Vet Sciences, James Cook University},
            addressline={James Cook Dr}, 
            city={Townsville},
            postcode={4811}, 
            state={QLD},
            country={Australia}}

\begin{abstract}
\paragraph{Introduction}
\noindent Advances in wearable sensor technology have enabled the collection of biomarkers that may correlate with levels of elevated stress. While significant research has been done in this domain, specifically in using machine learning to detect elevated levels of stress, the challenge of producing a machine learning model capable of generalizing well for use on new, unseen data remain. Acute stress response has both subjective, psychological and objectively measurable, biological components that can be expressed differently from person to person, further complicating the development of a generic stress measurement model. Another challenge is the lack of large, publicly available datasets labeled for stress response that can be used to develop robust machine learning models. In this paper, we first investigate the generalization ability of models built on datasets containing a small number of subjects, recorded in single study protocols. Next, we propose and evaluate methods combining these datasets into a single, large dataset to study the generalization capability of machine learning models built on larger datasets. Finally, we propose and evaluate the use of ensemble techniques by combining gradient boosting with an artificial neural network to measure predictive power on new, unseen data. 
In favor of reproducible research and to assist the community advance the filed, we make all our experimental data and code publicly available through Github at https://github.com/xalentis/Stress. This paper's in-depth study of machine learning model generalization for stress detection provides an important foundation for the further study of stress response measurement using sensor biomarkers, recorded with wearable technologies. 

\paragraph{Methods}
\noindent Sensor biomarker data from six public datasets were utilized in this study. Exploratory data analysis was performed to understand the physiological variance between study subjects, and the complexity it introduces in building machine learning models capable of detecting elevated levels of stress on new, unseen data. To test model generalization, we developed a gradient boosting model trained on one dataset (SWELL), and tested its predictive power on two datasets previously used in other studies (WESAD, NEURO). Next, we merged four small datasets, i.e. (SWELL, NEURO, WESAD, UBFC-Phys), to provide a combined total of 99 subjects, and applied feature engineering to generate additional features utilizing statistical summaries, with sliding windows of 25 seconds. We name this large dataset, StressData. In addition, we utilized random sampling on StressData combined with another dataset (EXAM) to build a larger training dataset consisting of 200 synthesized subjects, which we name SynthesizedStressData. Finally, we developed an ensemble model that combines our gradient boosting model with an artificial neural network, and tested it using Leave-One-Subject-Out (LOSO) validation, and on two additional, unseen publicly available stress biomarker datasets (WESAD and Toadstool).

\paragraph{Results}
Our results show that previous models built on datasets containing a small number (\textless 50) of subjects, recorded in single study protocols, cannot generalize well to new, unseen datasets. Our presented  methodology for generating a large, synthesized training dataset by utilizing random sampling to construct scenarios closely aligned with experimental conditions demonstrate significant benefits. When combined with feature-engineering and ensemble learning, our method delivers a robust stress measurement system capable of achieving 85\% predictive accuracy on new, unseen validation data, achieving a 25\% performance improvement over single models trained on small datasets. The resulting model can be used as both a classification or regression predictor for estimating the level of perceived stress, when applied on specific sensor biomarkers recorded using a wearable device, while further allowing researchers to construct large, varied datasets for training machine learning models that closely emulate their exact experimental conditions.

\paragraph{Conclusion}
Models trained on small, single study protocol datasets do not generalize well for use on new, unseen data and lack statistical power. Machine learning models trained on a dataset containing a larger number of varied study subjects capture physiological variance better, resulting in more robust stress detection. Feature-engineering assists in capturing these physiological variance, and this is further improved by utilizing ensemble techniques by combining the predictive power of different machine learning models, each capable of learning unique signals contained within the data. While there is a general lack of large, labeled public datasets that can be utilized for training machine learning models capable of accurately measuring levels of acute stress, random sampling techniques can successfully be applied to construct larger, varied datasets from these smaller sample datasets, for building robust machine learning models. 

\end{abstract}


\begin{highlights}
\item We show that existing research using machine learning techniques have utilized small datasets that fail to generalize on new, unseen stress biomarker data.
\item We propose a synthesizing method and produce a new dataset by engineering data from multiple small public datasets into a single larger dataset, and show that it results in high statistical power and better captures data variation for training more robust machine learning models.
\item We propose an ensemble approach for combining multiple machine learning models and demonstrate 25\% increased predictive performance over singular models.
\item In favor of reproducible research and to advance the field, our new dataset and all our codes are made publicly available. 
\end{highlights}

\begin{keyword}
Stress \sep Wearable sensor \sep Empatica E4 \sep Machine learning
\PACS 07.05.Mh \sep 87.85.fk
\MSC 68T01 \sep 92C99
\end{keyword}

\end{frontmatter}


\section{Introduction}

\noindent Stress can be defined as any type of change that causes physical, emotional or psychological strain. Such change in the environment elicits the activation of a cascade of biological responses (stress response) in the brain and in the body \cite{McEwen1998}. The stress response serves an important evolutionary role of helping the adaptation of the organism to the dynamically changing external and internal environment. This is achieved through mobilization of energy and its appropriate redistribution to organs that most immediately serve the adaptational response. In this sense the biological stress response is adaptive and beneficial in the short term. The risk lies with the fact that the biological systems and molecules sub-serving the stress response exert considerable effects and bio-energetic demands on the organism. If not properly shut down or controlled by the body’s feedback mechanisms, the long-term exposure to stress will have detrimental effects. In humans this may include an increased risk to develop metabolic, cardiovascular and mental disorders, resulting in significantly compromised quality of life and shortened life expectancy \cite{McEwen1993, McEwen2006}.\\

\noindent Wearable devices for personal health monitoring have increased significantly in technical sophistication and are capable of measuring a wide variety of physiological signals. Continuous measurement of these signals using wearables enable researchers to record and extract useful information to detect and monitor a variety of potential health-related events, including stress.\\

\noindent Despite limitations with battery time and number of available sensors in certain models, compared to controlled laboratory measurement devices, wearables are non-intrusive and easier to use. This ease has facilitated many experiments using wearables \cite{Sriramprakash2017, Limas2018, Schmidt2018, Nkurikiyeyezu2019,Eskandar2020,Siirtola2020, Indikawati2020g, Li2020, Iqbal2021g, Liapis2021, Ninh2021,Khan2022}, and predominantly utilizing Empatica's latest E4 device \cite{Empatica2022}, which have yielded a number of well-studied public datasets \cite{Kraaij2015,Birjandtalab2016,Schmidt2018,Haouij2018,Svoren2020,Sabour2021, Amin2022, Hosseini2022}. Table \ref{tab:datareviewed} provides a summary of the datasets which were considered in this study. The table also shows details about these datasets including their number of subjects, available biomarkers, their wearable devices, and labeling strategies, which will be discussed in more details.\\

\begin{sidewaystable}
\setlength\tabcolsep{1pt}
\centering
\caption{\label{tab:datareviewed}Summary of public wearable device stress-related datasets used in this study.}
\resizebox{\textwidth}{!}{
\begin{tabular}{p{8cm}cccccLLL}
\hline\hline
\textbf{Dataset}                                & \multicolumn{1}{l}{\textbf{Year |}} & \multicolumn{1}{l}{\textbf{Subjects |}} & \multicolumn{1}{l}{\textbf{Female |}} & \multicolumn{1}{l}{\textbf{Male |}} & \textbf{Duration |} & \textbf{Biomarkers |}                          & \textbf{Devices |}                                                             & \textbf{Labeling/Scoring}                                       \\
\hline
SWELL                                           & 2014                              & 25                                    & 8                                   & 17                                & 138 min           & EDA, HRV, ECG                                & Facial expression, body postures, Mobi                                   & Periodic: Neutral, Time Pressure, Interruptions                           \\
\rowcolor[rgb]{0.753,0.753,0.753} Neurological Status (NEURO)                             & 2017                              & 20                                    &                &              & 31 min            & ACC,EDA, TEMP, HR, SPO2                      & Empatica E4                                                                  & Periodic: Relax, Physical Stress, Emotional Stress, Relax, Emotional Stress, Relax \\

WESAD         & 2018                              & 15                                    & 3                                   & 12                                & 120 min           & ACC, EDA, BVP, IBI, HR, TEMP, ECG, EMG, RESP & RespiBAN, Empatica E4                                                        & Periodic: Preparation, Baseline, Amusement, Stress, Meditation, Recovery  \\
\rowcolor[rgb]{0.753,0.753,0.753} AffectiveROAD & 2018                              & 10                                    & 5                                   & 5                                 & 118 min           & EDA, HR, TEMP                                & Empatica and Zephyr BioHarness 3.0 chest belt                                & Scored by observer                  \\
Toadstool                                       & 2020                              & 10                                    & 5                                   & 5                                 & 50 min            & ACC, EDA, BVP, IBI, HR, TEMP                 & Empatica E4                                                                  & Periodic: Game play under time pressure                                                         \\ \rowcolor[rgb]{0.753,0.753,0.753} UBFC-Phys                                        & 2021                              & 56                                    & 46                                  & 10                                & 20 min           & EDA, BVP            & Empatica E4 & Self-report                        \\
A Wearable Exam Stress Dataset for Predicting Cognitive Performance in Real-World Settings                                       & 2022                              & 10                                    & 2                                   & 8                                 & 180 min            & EDA, HR, BVP, TEMP, IBI, ACC                 & Empatica E4                                                                  & Periodic                                                         \\ 
\rowcolor[rgb]{0.753,0.753,0.753} A multimodal sensor dataset for continuous stress detection of nurses in a hospital                                        & 2022                              & 15                                    & 15                                  & 0                                & Varying           & EDA, HR, ST, BVP, ACC, IBI            & Empatica E4 & Self-report                        \\
\hline\hline
\end{tabular}
}
\end{sidewaystable}

\noindent Currently, stress-related and personalized questionnaires are mainly used to measure or score (label) stress in real-life and outside of a laboratory context. However, this technique does not allow for continuous monitoring, and often suffers from bias such as demand effects, response and memory biases. Technology offers a solution, by combining the large quantities of sensor-based physiological data recorded using wearable devices with the use of machine learning techniques, and specifically for the purpose of measuring stress, supervised machine learning techniques. In supervised learning, models are trained using data that is accurately labeled for the response you are predicting for; in the context of this paper, the labeling would be for elevated levels of stress with labels as binary yes/no indicators or a numeric scale to indicate stress level, generally a range between 0 (no perceived stress) and 1 (maximum perceived stress).\\

\noindent As shown in  Table \ref{tab:datareviewed}, the datasets included in this study were labeled using one of two methods: (i) periodic \cite{Kraaij2015, Birjandtalab2016, Schmidt2018, Svoren2020, Amin2022}, where specific time frames during the experiment were either labeled as stressed or non-stressed, while the test subject was placed under that perceived condition (a stressful test or action, or non-stressed, restful period), or (ii) scored as experiencing stress or no stress during a particular period, either by completing a self-scoring evaluation \cite{Sabour2021, Hosseini2022}, or by an observer \cite{Haouij2018} who perceived a level of stress by observing the emotional reaction of the subject during that period.\\

\noindent When applying machine learning to the task of measuring stress on a biomarker dataset, a robust model should to be able to generalize well from the input (training) data for any new data from within the problem domain (validation data). Generalization in this context, therefore, refers to how well a trained machine learning model can perform on unseen data, i.e., data not included when initially training the model. In order to generate such a model, the training dataset should be sufficiently large and diverse. Within the context of this paper, this implies recording data samples under varying experimental conditions, and across demographics. Variance in the context of machine learning relates to the variety of predictions made by the model, while Bias refers to the distance of the predictions from the actual (true) values. A highly-biased model implies its predicted values are far from the actual, true values. A generalized model offers the best trade-off between bias and variance, thereby delivering the best predictive performance.\\

\noindent Power analysis assists researchers in determining the smallest sample size suitable to detect the effect of a given experiment at a desired level of significance, as collecting larger samples are likely costlier and much harder. One of the recurrent questions psychology researchers ask is: \emph{"What is the minimum number of participants I must test?"} \cite{Brysbaert2019}. The high number of participants required for an 80\% powered study often surprises cognitive psychologists, because in their experience, replicable research can be done with a smaller number. Given that an effect size of \emph{d = 0.4} is a good first estimate of the smallest effect size of interest in psychological research \cite{Brysbaert2019}, over 50 subjects will be required for a simple comparison of two within-participants conditions to run a study with 80\% power. Of the datasets included in this study, only UBFC-Phys \cite{Sabour2021} contains biomarker data for more than 50 subjects.\\

\noindent In this work, we investigate whether models built on datasets with a small number of study participants (\textless 50) are capable of generalizing well for use on new, unseen data, thereby showing transferability and generalizability. We further propose to improve the statistical power of training datasets by merging a number of small datasets to create a larger dataset named StressData (with \textgreater 90 subjects) to improve machine learning model generalizability to new unseen datasets. We also propose a new method of synthesizing data from a number of available small datasets to make a larger  balanced dataset. We call this new dataset SynthesizedStressData and show that it can significantly improve generalizability of various models to new unseen data. Finally, we propose to combine the predictive power of two unique machine learning algorithms using ensemble methods, and investigate whether our proposed approach offers improved predictive power over singular models in the literature.

\section{Related Work}

\noindent A growing number of studies have been conducted in recent years with the aim of building robust stress detection and measurement machine learning models. Few studies have further examined the reproducibility and generalizability of results previously reported. Mishra \emph{et al.} \cite{mishra2020} took the first steps towards evaluating the performance of models built using data from one study, and testing their performance on data from other studies. Reproducing the results achieved in previous studies rely on the availability of the stress biomarker data, and most often this was not the case.\\

\noindent Where experimental data was made available and the studies utilized Electrodermal Activity (EDA) and Heart Rate (HR) biomarkers, the reported results and associated models and datasets were reviewed as listed in Table \ref{tab:modelsreviewed}. The publicly available WESAD \cite{Schmidt2018} dataset containing sensor biomarker data for 15 subjects was the most commonly used \cite{Schmidt2018, Nkurikiyeyezu2019, Indikawati2020g, Li2020, Iqbal2021g, Liapis2021, Ninh2021}. The NEURO \cite{Birjandtalab2016} dataset was utilized in experiments by Jiménez-Limas \emph{et al.} \cite{Limas2018} and Eskandar \emph{et al.} \cite{Eskandar2020}, while Siirtola \emph{et al.} \cite{Siirtola2020} and Ninh \emph{et al.} \cite{Ninh2021} utilized the AffectiveROAD dataset containing sensor biomarker data for 9 subjects. Sriramprakash \emph{et al} \cite{Sriramprakash2017} and Khan \cite{Khan2022} utilized the SWELL \cite{Kraaij2015} dataset containing sensor biomarker data for 25 subjects, while Nkurikiyeyezu \emph{et al.} \cite{Nkurikiyeyezu2019} utilized both the WESAD and SWELL datasets.\\

\noindent Accuracy metrics reported ranged between 81.13\% \cite{Ninh2021} and 99.80\% \cite{Li2020}. LOSO cross-validation were utilized in three experiments \cite{Schmidt2018, Siirtola2020, Khan2022}. Indikawati \emph{et al.} \cite{Indikawati2020g} used a 60/40 train/validation split for validating their model performance, while Li \emph{et al.} \cite{Li2020} used a 70/30 train/validation split. Jiménez-Limas \emph{et al.} \cite{Limas2018} and Eskandar \emph{et al.} \cite{Eskandar2020} both used an 80/20 split for their experiments utilizing the NEURO dataset. The remaining studies utilized n-Fold cross-validation.\\

\begin{sidewaystable}
\centering
\caption{\label{tab:modelsreviewed}Summary of related works reviewed.}
\renewcommand{\arraystretch}{1.5}
\resizebox{\textwidth}{!}{
\begin{tabular}{Lcccccccc}
\hline\hline
\textbf{Paper}                                                                                                                                                               & \multicolumn{1}{l}{\textbf{Year}} & \textbf{Model}                & \textbf{Dataset} & \multicolumn{1}{l}{\textbf{Accuracy}} & \multicolumn{1}{l}{\textbf{Subjects}} & \multicolumn{1}{l}{\textbf{Features}} & \textbf{Cross Validation}     & \textbf{Window}    \\
\hline
\rowcolor[rgb]{0.753,0.753,0.753} Stress Detection in Working People \cite{Sriramprakash2017}                                                                    & 2017                     & SVM                          & SWELL                & 92.75\%                             & 25                           & 17                           & 10-Fold          & 60s                \\
Feature selection for stress level classification into a physiologycal signals set \cite{Limas2018}
& 2018                     & Linear Regression & NEURO                & 81.38\%                             & 20                           & 7                           & 80/20 Split             & 5m     \\
\rowcolor[rgb]{0.753,0.753,0.753} Introducing WESAD, a Multimodal Dataset for Wearable Stress and Affect Detection \cite{Schmidt2018}                            & 2018                     & Random Forest, LDA, AdaBoost & WESAD                & 93.00\%                             & 15                           & 82                           & LOSO             & 0.25s, 5s, 60s     \\
he Effect of Person-Specific Biometrics in Improving Generic Stress Predictive Models \cite{Nkurikiyeyezu2019}                & 2019                     & Random Forest, ExtraTrees    & WESAD, SWELL         & 93.90\%                             & \multicolumn{1}{l}{15, 25}   & 94                           & 10-Fold          & 5min, 10min        \\
\rowcolor[rgb]{0.753,0.753,0.753} Using Deep Learning for Assessment of Workers' Stress and Overload \cite{Eskandar2020}  & 2020                     & Neural Network         & NEURO        & 85\%                             & 20                            & 4                          & 80/20 Split              & 20s  \\
Comparison of Regression and Classification Models for User-Independent and Personal Stress Detection \cite{Siirtola2020}      & 2020                     & Bagged tree ensemble         & AffectiveROAD        & 82.30\%                             & 9                            & 119                          & LOSO             & 60s, 0.5s overlap  \\
\rowcolor[rgb]{0.753,0.753,0.753} Stress Detection from Multimodal Wearable Sensor Data \cite{Indikawati2020g}                                                   & 2020                     & Random Forest                & WESAD                & 92.00\%                             & 15                           & 4                            & 60/40 Split      & 0.25s              \\
Stress detection using deep neural networks \cite{Li2020}                                                                      & 2020                     & Neural Network               & WESAD                & 99.80\%                             & 15                           & 8                            & 70/30 Split      & 5s                 \\
\rowcolor[rgb]{0.753,0.753,0.753} A Sensitivity Analysis of Biophysiological Responses of Stress for Wearable Sensors in Connected Health \cite{Iqbal2021g}      & 2021                     & Logistic Regression          & WESAD                & 85.71\%                             & 14                           & 5                            & 14-Fold          & 60s                \\
Advancing Stress Detection Methodology with Deep Learning Techniques \cite{Liapis2021}                                         & 2021                     & SVM                        & WESAD                & 93.20\%                             & 15                           & 36                           & 5-Fold           &                    \\
\rowcolor[rgb]{0.753,0.753,0.753} Analysing the Performance of Stress Detection Models on Consumer-Grade Wearable Devices \cite{Ninh2021}                        & 2021                     & SVM                          & WESAD, AffectiveROAD & \multicolumn{1}{l}{87.5\%, 81.13\%} & \multicolumn{1}{l}{15, 9}    & 1                            &                  & 60s, 30s           \\
Semi-Supervised Generative Adversarial Network for Stress Detection Using Partially Labeled Physiological Data \cite{Khan2022} & 2022                     & Neural Network               & SWELL                & \multicolumn{1}{l}{ 90.00\%}        & 25                           & 30                           & LOSO             & 60s  
    \\
\hline\hline
\end{tabular}
}
\end{sidewaystable}

\noindent Feature-engineering techniques were employed in all the models reviewed apart from Indikawati \emph{et al.} \cite{Indikawati2020g}, where Accelerometer (ACC), Electrodermal Activity (EDA), Blood Volume Pulse (BVP) and temperature (TEMP) signals were directly used, and Eskandar \emph{et al.} \cite{Eskandar2020} who utilized HR, EDA, TEMP and arterial
oxygen levels (SpO2). Iqbal \emph{et al.} \cite{Iqbal2021g} used a chest-worn RespiBAN device with the Empatica E4 and included biomarker data for RR-interval (RRI) and respiratory rate (RspR). Four of the models \cite{Schmidt2018, Nkurikiyeyezu2019, Siirtola2020, Indikawati2020g} employed tree-based machine learning models while Iqbal \emph{et al.} \cite{Iqbal2021g} and Jiménez-Limas \emph{et al.} \cite{Limas2018} utilized Logistic Regression and Linear Regression, respectively. Support Vector Machines (SVM) were utilized in experiments by Sriramprakash \emph{et al}  \cite{Sriramprakash2017}, Liapis \emph{et al.} \cite{Liapis2021} and Ninh \emph{et al.} \cite{Ninh2021}. Artificial neural network models were utilized in the remaining three studies \cite{Eskandar2020, Li2020, Khan2022} and reported the highest predictive accuracy across all results, followed by tree-based models. The most common feature-engineering technique employed consisted of generating new features from statistical summaries of biomarker data, based on grouped sliding windows ranging between 0.25 seconds to 60 seconds. Nkurikiyeyezu \emph{et al.} \cite{Nkurikiyeyezu2019} and Jiménez-Limas \emph{et al.} \cite{Limas2018} employed larger sliding windows of 5 minutes and 10 minutes, respectively.\\

\noindent In the related works reviewed, experimentation was performed on a single dataset, most notably WESAD, with a limited number of subject sensor biomarker data (15 subjects). Only Nkurikiyeyezu \emph{et al.} \cite{Nkurikiyeyezu2019} validated their experimentation results by using two unique datasets, WESAD and SWELL (15 and 25 subjects, respectively). As noted previously, Brysbaert \emph{et al.} \cite{Brysbaert2019} recommended over 50 subjects to achieve statistical power of over 80\%, which brings into question whether models built on datasets containing less than 50 subjects produce results that can be considered statistically significant, and whether these models generalize well to predict accurately on new, unseen data.\\

\noindent To address this question, in this paper we will test the generalizability of machine learning models built on one, small dataset by validating these models on different, unseen datasets. The ability of machine learning models to generalize well is crucial to providing a technology solution that can be utilized and applied outside of experimental conditions, thereby providing a step forward towards enabling health practitioners to better use sensor biomarker data recorded using wearable devices.\\

\noindent In addition, in order to fill the gap of lack of substantially large public datasets available for stress research, we propose to merge four small public stress biomarker datasets into a single larger dataset (StressData), representing a total of 99 test subjects, thereby providing statistical power of over 85\%. Although we show that the improved statistical power of a larger number of subjects within a training dataset leads to improved predictive performance, a challenge with the merging approach is the resulting class imbalance created due to the varying experimental protocols utilized during biomarker recording. To address this challenge, we propose using random sampling of StressData segments to generate a 200-subject synthetic training dataset, which we name SynthetizedStressData. Finally, we build two unique machine learning models utilizing the well-known XGBoost (XGB) gradient boosting algorithm and an artificial neural network (ANN), then combine their predictive power using ensemble methods to deliver a robust stress detection system. We show that while our ensemble model trained on StressData leads to small improvements, utilizing our SynthesizedStressData dataset results in a predictive performance improvement of over 25\%, compared to another model trained on a small dataset with low statistical power.

\section{Methods}
\subsection{Datasets}

\noindent Based on the prior experiments and reviewed literature (Table \ref{tab:modelsreviewed}), publicly available datasets including WESAD \cite{Schmidt2018} and SWELL \cite{Kraaij2015} were utilized in this study. Additionally, since these datasets all included  Empatica E4 sensor biomarker data, the Toadstool \cite{Svoren2020}, UBFC-Phys \cite{Sabour2021}, Non-EEG Dataset for Assessment of Neurological Status (NEURO) \cite{Birjandtalab2016}, Wearable Exam Stress Dataset (EXAM) \cite{Amin2022}, AffectiveROAD \cite{Haouij2018} and Multimodal Sensor Dataset for Continuous Stress Detection of Nurses in a Hospital \cite{Hosseini2022} public datasets, which also are collected using Empatica E4, were considered. Table \ref{tab:datareviewed} provides a summary of all  datasets considered. The AffectiveROAD dataset was excluded from our experimentation due to its labeling protocol. In this dataset, subjects were self- and observer-scored on a scale that limited conversion to a binary stressed/non-stressed indicator for model training and evaluation purposes. Subjects were scored while driving in inner city and highway scenarios, and it is not clear which scenario under normal circumstances would be considered more or less stressful.\\

\noindent Additionally, the Multimodal Sensor Dataset for Continuous Stress Detection of Nurses in a Hospital \cite{Hosseini2022} was not included in our experimentation for similar reasons. In this dataset, thirteen potential scenarios could be marked by the subject as stressful by using the Empatica E4 event marker button. We found a substantial number of marked sections that exceeded the expected duration of the event, with a further number of short events with no potential for a cooling down period to separate the perceived stressful event from a subject baseline (non-stressed).\\

\noindent The SWELL, WESAD, NEURO and UBFC-Phys datasets were specifically labeled for stress, either using self-report scoring (UBFC-Phys) or periodic (SWELL, NEURO, WESAD, EXAM), where stressful situations were applied during specific time periods of the experiment. The Toadstool dataset contained sensor biomarker data recorded during periods of game-play where stress may have been perceived as high, but was not labeled. However, it was included to validate the XGB, ANN and ensemble models developed as part of this study. The EXAM dataset was also included for some of our experiments. This dataset was recorded during single sessions of 10 students writing mid-term and final exams, with exam scores provided, but no specific stress-based scoring included.\\

\begin{figure}[!h]
\centering
\fbox{\includegraphics[width=\textwidth]{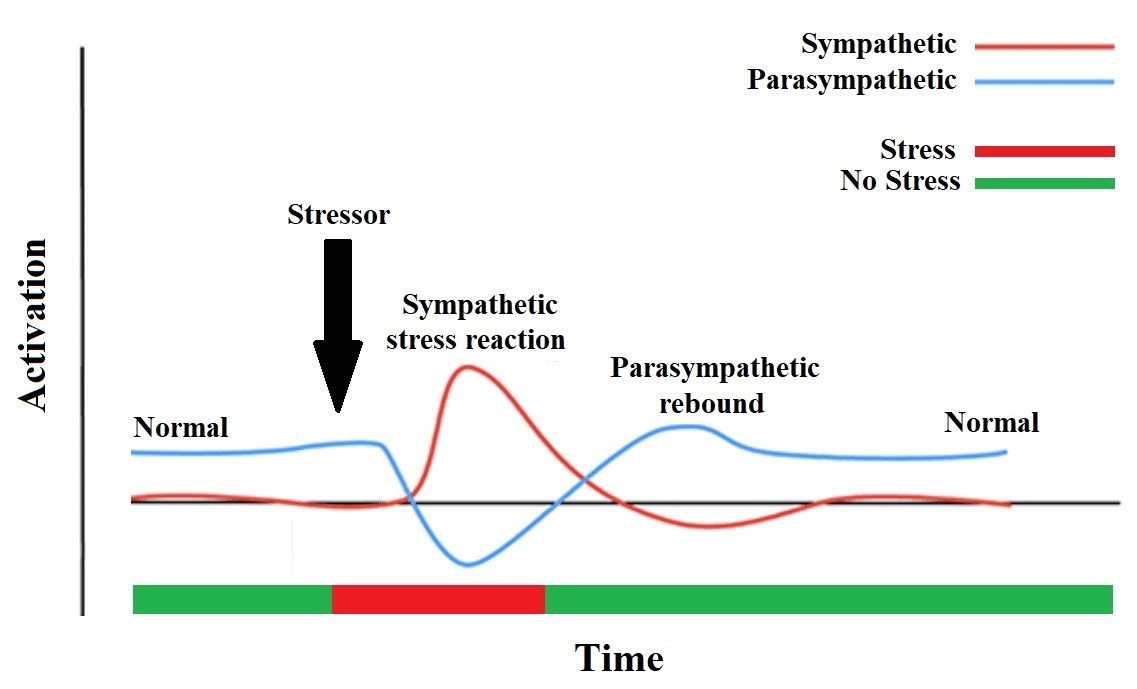}}
\caption{\label{fig:figure1}Stress response with binary labeling over time.}%
\end{figure}

\noindent As stress response is not an instantaneous physiological reaction, overlap may occur between the labeled periods within the datasets and the biomarker data, as shown in Figure \ref{fig:figure1}. This overlap can introduce miss-labeled biomarker data into a training set when building machine learning models and will likely further produce onerous predictions in new datasets during inference. It is therefor important to remove this overlap by excluding a short timespan prior to the stressor phase, and likely a longer period when the stressor phase completes to allow for a cool-down period during model training. Additionally, any periods labeled not clearly defined as stressed or non-stressed (baseline) should likely be excluded. In this paper, the following exclusions were applied to the datasets utilized:

\begin{itemize}
  \item WESAD - Period marked as Meditation was excluded.
  \item NEURO - 2 Minute period after stress event was excluded.
  \item SWELL - 2 Minute period after stress event was excluded.
  \item UBFC-Phys - 5 Minute introductory period was excluded, 3 minute period after stress event was excluded.
  \item EXAM - No exclusions made.
\end{itemize}

\noindent Of the biophysical and biochemical markers normally used to measure stress \cite{Iqbal2021g}, in this study we opted to focus on Electrodermal Activity (EDA), and Heart Rate (HR). Previous works have included Accelerometer (ACC) bio-marker data as a feature, however considering that at least one of the datasets (Toadstool) involved substantial subject physical movement as part of their experimental setup, this signal was excluded. Temperature (TEMP) and Skin Temperature (ST) was further excluded due to being missing in two additional datasets (SWELL, UBFC-Phys).\\

\noindent While not all datasets included an HR biomarker, this can be algorithmically derived using Heart Rate Variation (HRV) or Blood Volume Pulse (BVP), and where required was generated using a Python script utilizing the BioSPPy library \cite{BioSPPy}.\\

\subsection{Dataset analyses}
\noindent The statistical programming language R \cite{RSoftware} version 4.0.4 was used for analysis in this study. The Empatica-R library \cite{EmpaticaR} was additionally utilized to read the raw Empatica E4 wearable device data into an R data table for processing. Next, sensor signals were converted to the same sampling frequency, as the Empatica E4 samples the BVP sensor data at 64 Hz, the EDA sensor data at 4 Hz, and HR sensor data in spans of 10 seconds. Stress labels within the datasets were then standardized to binary indicators with non-stressed set to zero (0) and stressed set to one (1).\\

\noindent Initial data exploration was performed on only the NEURO, WESAD, SWELL and UBFC-Phys datasets, considering the WESAD and SWELL datasets were utilized in prior work as listed in Table \ref{tab:modelsreviewed}, and all four contained at least the EDA and HR biomarker data for a substantial percentage of the observations available. Not all biomarker sensor data were available for all subjects within the SWELL and WESAD datasets, and in these cases the subjects were excluded. This resulted in a total of 9 subjects from the SWELL dataset being included (of 25) containing a total of 157,739 observations, and 14 of the 15 subjects from the WESAD dataset, containing a total of 26,385 observations.\\

\noindent Statistical summaries were produced for each dataset, as detailed in Table \ref{tab:datastats}, indicating substantial class imbalance across the datasets, with the NEURO and UBFC-Phys datasets containing 70.1\% and 65.8\% labels marked as stressed, respectively. In contrast, WESAD and SWELL have labels marked as non-stressed in 63.6\% and 71.4\% of total observations. The Table also shows the mean and Standard Deviation (SD) of the biomarkers across the four datasets. Substantial variation is further noted across each dataset for both the EDA and HR biomarkers.

\begin{table}
\setlength\tabcolsep{4pt}
\centering
\caption{\label{tab:datastats}Statistical summary of included public wearable device stress-related datasets.}
\resizebox{\textwidth}{!}{
\begin{tabular}{ccccccc}
\hline\hline
\textbf{Dataset}                        & \multicolumn{1}{c}{\textbf{EDA}}    & \multicolumn{1}{c}{\textbf{EDA}}  & \multicolumn{1}{c}{\textbf{HR}}     & \multicolumn{1}{c}{\textbf{HR}}   & \multicolumn{1}{l}{\textbf{Non-Stressed}} & \multicolumn{1}{l}{\textbf{Stressed}}  \\
                                        & \multicolumn{1}{c}{\textbf{(Mean)}} & \multicolumn{1}{c}{\textbf{(SD)}} & \multicolumn{1}{c}{\textbf{(Mean)}} & \multicolumn{1}{c}{\textbf{(SD)}} & \multicolumn{1}{l}{}                      & \multicolumn{1}{l}{}                   \\
\rowcolor[rgb]{0.753,0.753,0.753} NEURO & 1.7                                 & 2.1                               & 75.4                                & 13.4                              & 29.90\%                                   & 70.10\%                                \\
WESAD                                   & 2.2                                 & 2.7                               & 76.5                                & 10.5                              & 63.60\%                                   & 36.40\%                                \\
\rowcolor[rgb]{0.753,0.753,0.753} SWELL & 1                                   & 0.2                               & 75.8                                & 10.1                              & 71.40\%                                   & 28.60\%                                \\
UBFC-Phys                                    & 1.4                                 & 2                                 & 92.2                                & 16                                & 34.20\%                                   & 65.80\%                               

\\
\hline\hline
\end{tabular}
}
\end{table}

\begin{figure}[h]
\centering
\fbox{\includegraphics[width=\textwidth]{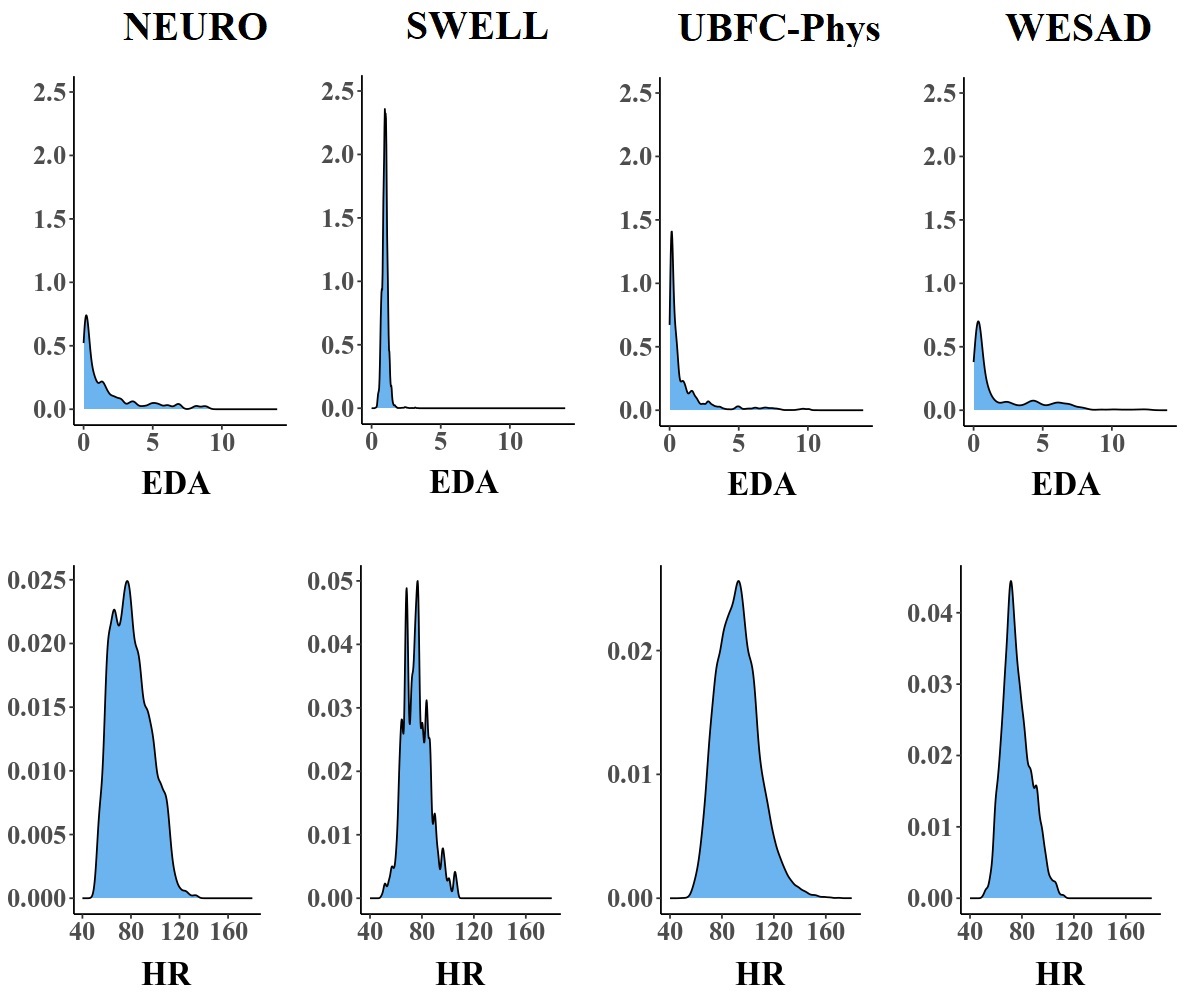}}
\caption{\label{fig:figure2}Biomarker histograms for NEURO, SWELL, UBFC-Phys and WESAD datasets.}%
\end{figure}

\noindent Histograms (Figure \ref{fig:figure2}) of each biomarker across the datasets show the EDA biomarker being significantly right-skewed. It should be noted that, the SWELL dataset contains on average 4\% or more observations than the NEURO, WESAD or UBFC-Phys datasets, while the UBFC-Phys dataset contains the largest number of individual subjects, but each recording segment being significantly shorter compared to the SWELL, NEURO and WESAD datasets.\\
 
\begin{figure}
\centering
\fbox{\includegraphics[width=\textwidth]{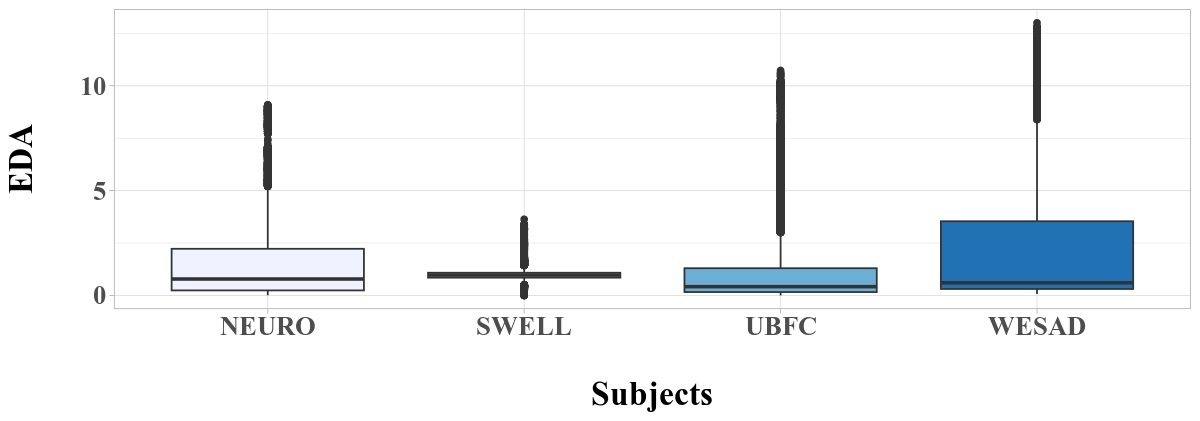}}
\caption{\label{fig:figure3}EDA biomarker variance across NEURO, SWELL, UBFC-Phys and WESAD dataset subjects.}%
\end{figure}

\begin{figure}
\centering
\fbox{\includegraphics[width=\textwidth]{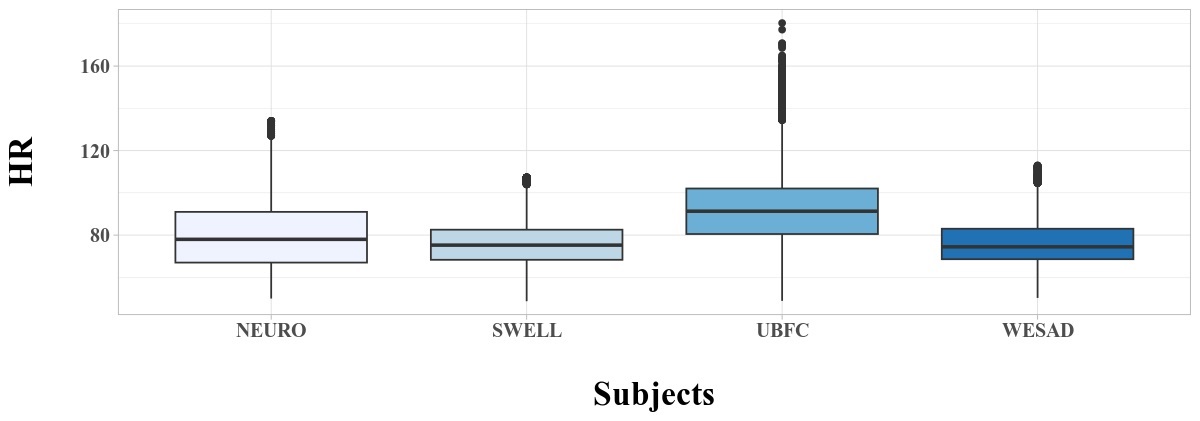}}
\caption{\label{fig:figure4}HR biomarker variance across NEURO, SWELL, UBFC-Phys and WESAD dataset subjects.}%
\end{figure}

\noindent Box plots highlight this variation further (Figure \ref{fig:figure3} and Figure \ref{fig:figure4}), with significant outliers occurring for the HR biomarker (Figure \ref{fig:figure4}) of the UBFC-Phys dataset.\\

\begin{figure}[!h]
\centering
\fbox{\includegraphics[width=\textwidth]{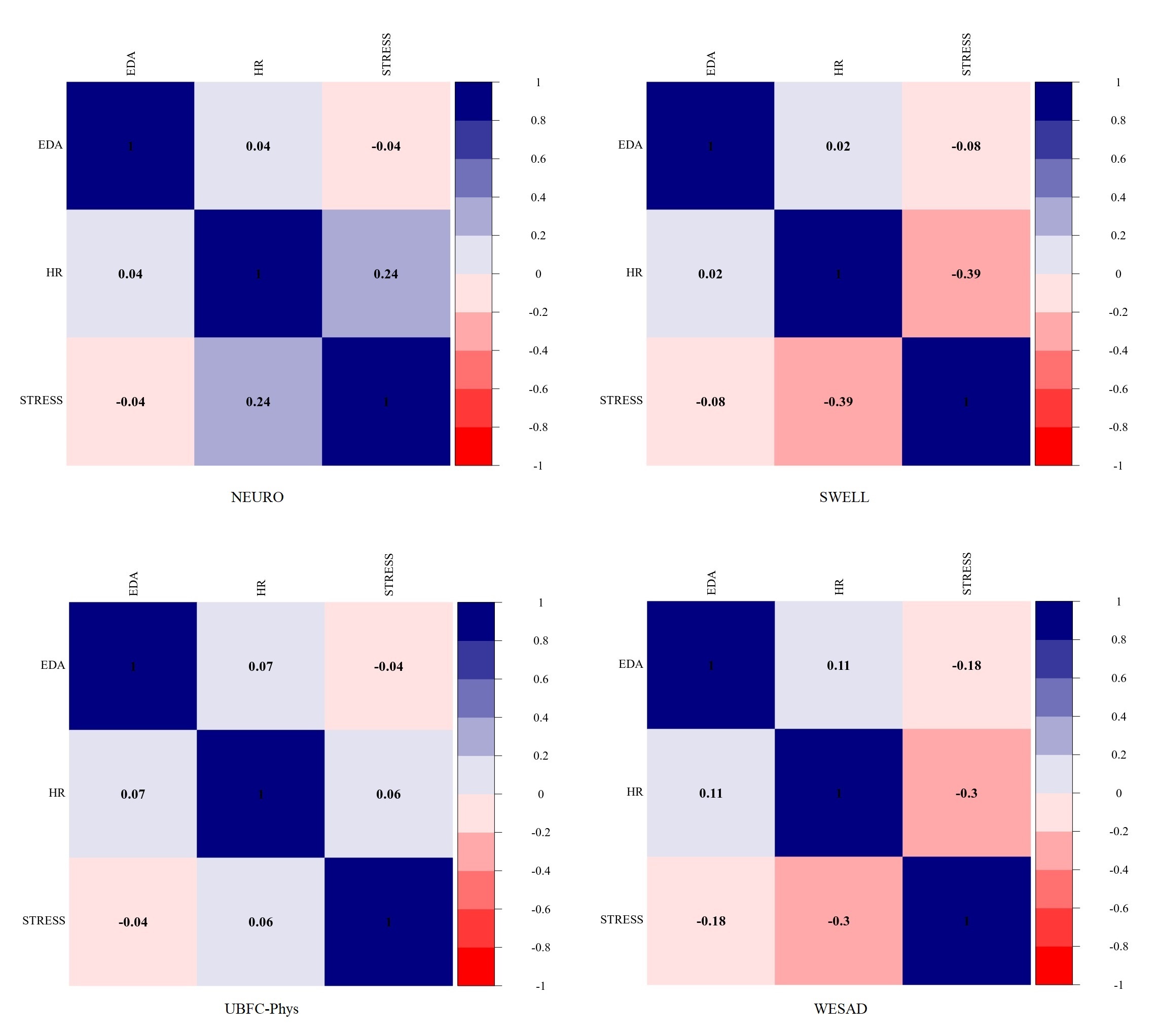}}
\caption{\label{fig:figure5}Correlation between biomarkers and stress metric of NEURO, SWELL, UBFC-Phys and WESAD datasets.}%
\end{figure}

\noindent In addition to box plots and statistical summaries, plots were generated for each of the four datasets to investigate correlation between the HR and EDA biomarkers and the labeled, binary stress metric (Figure \ref{fig:figure5}). Correlation was consistently weak across all the datasets, with the lowest correlation observed in the UBFC-Phys dataset. Within the NEURO dataset, we noted higher correlation between the binary stress label and the HR biomarker, while correlation was low between the binary stress label and the EDA biomarker. This correlation was slightly higher for the SWELL and WESAD biomarkers. This difference in correlation across stress biomarker datasets can affect transferability of experimental results, and further limit the ability of machine learning models trained on any of these individual datasets to generalize well for new, unseen data.
 
\subsection{Machine Learning Models} \label{sec:ML}

\noindent One of the most crucial steps in building machine learning models is the choice of algorithm. A common approach for optimal algorithm selection involves conducting exhaustive grid-searches using a wide variety of algorithms and approaches, and determining the best potential algorithm based on the results achieved. By examining the reported results of previous experiments and their associated algorithms (summarized in Table \ref{tab:modelsreviewed}), we find tree-based algorithms including Random Forest performing well compared to other algorithms, in line with findings reported by Mishra \emph{et al.} \cite{mishra2018}, who noted that Random Forest is often the optimal model when using EDA, HR and RRI biomarker data as input features for stress-related machine learning models.\\

\noindent Gradient Boosting algorithms, compared to Random Forest, offer improved efficiency and predictive performance, at the cost of being prone to over-fitting. Random Forest builds a number of decision trees independently, where each decision tree is a simple predictor with all results aggregated into a single result. The Gradient Boosting algorithm, however, is an ensemble of weak predictors, usually decision trees. Additionally, in Random Forest, the results of the decision trees are aggregated at the end of the process, while Gradient Boosting instead aggregates the results of each decision tree along the way to calculate the final result. Popular Gradient Boosting algorithms include LightGBM \cite{LightGBM}, CatBoost \cite{CatBoost} and XGBoost (XGB) \cite{XGBoost}.\\

\noindent For this study, XGB was selected as the primary machine learning algorithm due to its popularity, availability as an R package, and having been extensively utilized since its original publication in 2016 \cite{XGBoost}. XGB provides a highly efficient implementation of the stochastic gradient boosting algorithm and access to a suite of model hyperparameters, designed to provide control over the model training process.

\subsection{Model Generalization} \label{sec:modelgen}
\noindent To test the generalizability of machine learning models built on datasets containing sensor biomarker data of a small number of subjects, two initial experiments were performed, as follows:
\begin{enumerate}
  \item Train Random Forest, SVM and XGB on SWELL, test on NEURO and WESAD with no additional feature-engineering, using EDA and HR biomarkers only.
  \item Train Random Forest, SVM and XGB on SWELL, test on NEURO and WESAD and generate additional features using statistical summaries.\\
\end{enumerate}

\noindent The SWELL dataset contains the largest number of subjects and observations, nearly 6 times more than that of either WESAD or NEURO. Additionally, the SWELL dataset showed the lowest standard deviation for both HR and EDA biomarkers (Table \ref{tab:datastats}), therefore, it was selected for training. However, we also conducted experiments where either of the smaller WESAD or NEURO datasets were used for training. As expected, these experiments yielded poor results due to the low number of subjects and observations leading to a lack of statistical significance. These results and their code are made available through the paper’s public code repository (see Supplement6.R). \\ 

\noindent The UBFC-Phys dataset was excluded as it contained 56 subjects, substantially larger than WESAD, NEURO or SWELL, which were used in the previous works that were reviewed. As the stress metric within the datasets was binary, logistic regression was selected as the XGB learning objective. Due to the metric imbalance within each dataset, class balancing was performed by using the \emph{scale\textunderscore pos\textunderscore weight} parameter provided by XGB, which has the effect of weighing the balance of positive examples, relative to negative examples, when boosting decision trees. Random Forest and SVM models were included to create baseline and compare initial results to those reported in previous studies (Table \ref{tab:modelsreviewed}).\\

\noindent Optimal hyperparameters for the XGB algorithm were identified using a grid search with 10-Fold cross-validation, and this was repeated for each experiment, to ensure the most suitable parameters were utilized prior to model training. The XGB algorithm was run for 5,000 rounds with early-stopping set to 3 rounds. This ensures that training stops when the evaluation metric fails to improve after 3 rounds of training in order to prevent over-fitting. For the initial two experiments, training was automatically stopped after 93, and 87 rounds, respectively. For the Random Forest model, the number of trees were set to 200 \cite{Latinne2001}, while a radial kernel was selected for the SVM model with the \emph{cost} parameter set to 5, which were found to be optimal based on a number of trial experiments which were run with varying kernels and cost parameters ranging from 1 to 10.\\

\begin{figure}[h]
\centering
\fbox{\includegraphics[width=\textwidth]{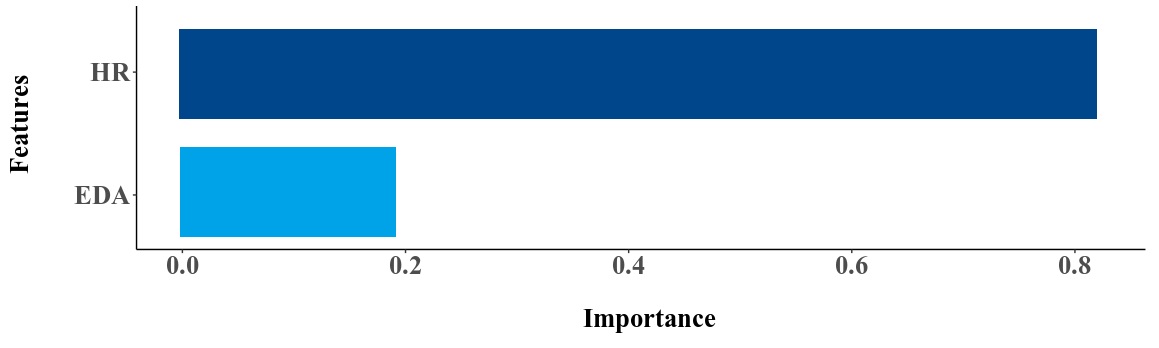}}
\caption{\label{fig:figure6}Feature importance for experiment 1, i.e. before biomarker feature engineering.}%
\end{figure}

\noindent Prior to performing experiments where feature-engineering was applied, we investigated the dominance of biomarkers and found HR feature as a more important biomarker, relative to EDA, as shown in Figure \ref{fig:figure6}.\\

\noindent Feature importance was determined using the built-in \emph{xgb.plot.importance} function provided by the XGBoost R package. Here, importance is calculated based on overall gain, where gain implies the relative contribution of the corresponding feature to the model, calculated by taking each feature's contribution for each tree in the model. A higher value of this metric when compared to another feature implies it is more important for generating a prediction.\\

\begin{figure}[h]
\centering
\fbox{\includegraphics[width=\textwidth]{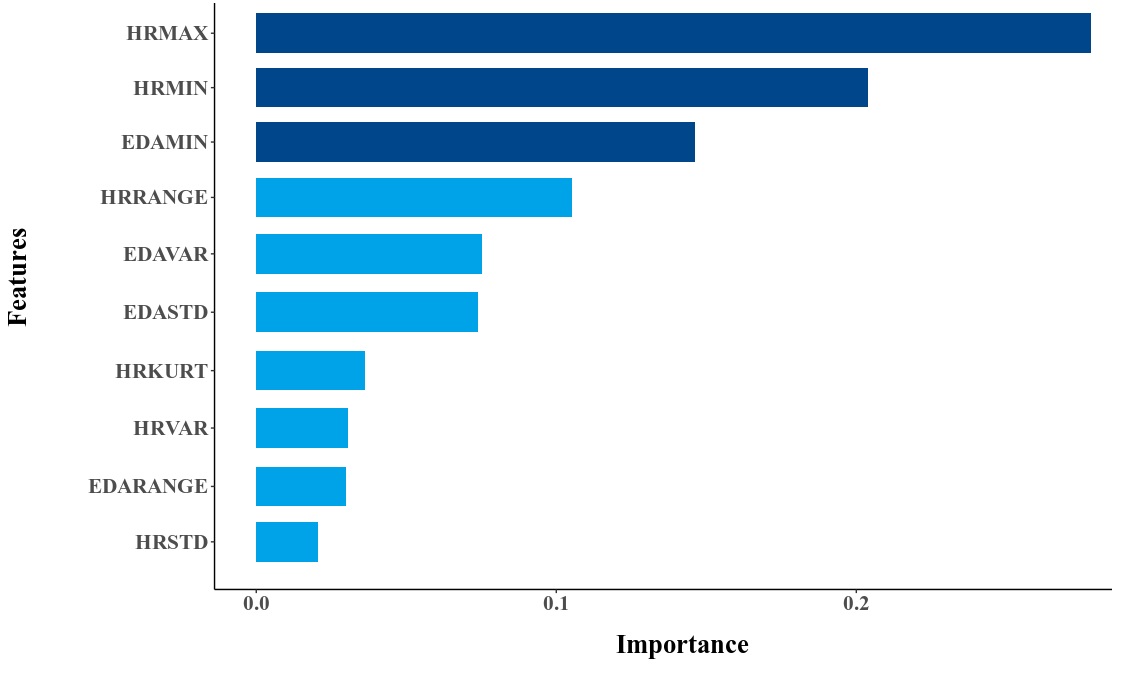}}
\caption{\label{fig:figure7}Feature importance for experiment 2, i.e. after biomarker feature engineering.}
\end{figure}

\noindent Next, an additional 22 features were generated by calculating statistical summaries using a tumbling window approach at 25-second intervals. Intervals between 0.25 seconds and 60 seconds are common in the literature (see Table \ref{tab:modelsreviewed}), and 25-second intervals were selected based on the highest level of correlation observed between the biomarkers and stress label, after a substantial amount of experimentation. Statistical summaries including the mean, median, max, min, standard deviation, variance, skewness and kurtosis of each biomarker were calculated as new features, with an additional feature for the covariance between the EDA and HR biomarker. The resulting variable importance after model training and evaluation is detailed in Figure \ref{fig:figure7} (top 10, in order of importance).\\

\noindent The HR biomarker and its derived features are again observed to be the dominant features, with EDA-based features serving a secondary role. These 10 features were utilized in all our experiments requiring feature engineering. The remaining 12 features were not utilized to prevent potential over-fitting to specific datasets, notably UBFC-Phys, with 56 subjects, and substantially shorter recording segments compared to the NEURO, WESAD and SWELL datasets.\\

\subsection{Merging Unique Stress Biomarker Datasets to Form StressData}

\noindent In order to increase the statistical power of the underlying smaller datasets, all four smaller datasets, i.e.  NEURO, WESAD, SWELL and UBFC-Phys were merged row-wise without randomization, to form the single larger dataset, StressData. This includes a total of 99 unique subjects resulting in more than 85\% statistical power, constituting a total of 244,399 observations, as follows:

\begin{itemize}
  \item All 20 subjects from the NEURO dataset.
  \item All 56 subjects from the UBFC-Phys dataset.
  \item 14 of the 15 WESAD dataset subjects contained sufficient biomarker data for inclusion.
  \item 9 of the 25 SWELL dataset subjects contained sufficient biomarker data for inclusion.
\end{itemize}

\noindent Two new experiments, numbered 3-4, were then performed on the StressData dataset, as follows:

\begin{enumerate}
\setcounter{enumi}{2}
  \item Train XGB on StressData using LOSO validation, with no feature-engineering, using EDA and HR biomarkers only.
  \item Train XGB on  StressData using LOSO validation, apply  feature-engineering (10 total features), as explained in subsection \ref{sec:modelgen}.
\end{enumerate}

\subsection{An Artificial Neural Network Model for Stress Detection}

\noindent Of the prior works reviewed, we found three instances of using Artificial Neural Networks (ANN). Li \emph{et al.} \cite{Li2020} utilized a Deep Convolutional Network (DCNN) consisting of three hidden layers containing 32 and 16 units, respectively. This was followed by a Sigmoid activation layer, with max pooling applied at the end of each convolution. Predictive accuracy using the WESAD dataset was reported as 99.80\%. Khan \cite{Khan2022} tested Long-Short Term Memory Network (LTSM),  Convolutional Neural Networks (CNN) and Bi-Directional Long-Short Term Memory networks, using both supervised and semi-supervised methods and found supervised models to perform better than semi-supervised models, with accuracy across predictions averaging higher than 90\% on the SWELL dataset. Eskandar \emph{et al.} \cite{Eskandar2020} utilized a LTSM model fed with features extracted by using a CNN with a reported accuracy of 85\% (F1 Score), noting difficulty on predicting the baseline relaxed state when using the NEURO dataset.\\

\noindent Based on recommendations from available literature \cite{Stathakis2009} and in consideration of the different experimental protocols and stressed/non-stressed period durations contained within the datasets utilized, we opted for a standard feed-forward architecture as a baseline model, rather than an LTSM model, which is more suitable for time-series data. Our proposed network consists of three layers. The input layer contained 10 neurons to receive the 10 input features. The hidden layer included 5 neurons (half the input features), connected to a final linear layer.\\

\noindent Mean Squared Error (MSE) was selected as the loss function, a frequently used measure of the differences between values predicted by a model and the expected values:

$$MSE = \sum_{i=1}^{N}(\hat{y_i}-y_i)^2$$

\noindent where N is the number of observations, $\hat{y}$ is the predicted value and \emph{y} is the expected value. Typically, binary and multi-class problems including XGBoost would utilize log-loss as a loss function instead:

$$LogLoss = - \frac{1}{N} \sum\limits_{i=1}^N [y_i \cdot log_e(\hat{y_i}) + (1-y_i) \cdot log_e(1-\hat{y_i}) ]$$

\noindent where N is the number of observations, log is the natural log, \emph{y} is the binary indicator (0 or 1), and $\hat{y}$ is the predicted  observation. However, we found MSE to outperform log-loss by a factor of 10\% for predictive accuracy, and therefor utilized MSE for the neural network model, while retaining the default log-loss for XGBoost. The precision, recall and F-score metrics were also calculated given the classification nature of the labeling of the SWELL, WESAD, NEURO and UBFC-Phys datasets.\\

\noindent The Rectified Linear Unit (ReLU) \cite{RELU} activation function was applied after both the input and hidden layers. Training was performed on a dual-GPU system using the Keras and Tensorflow libraries for a maximum of 200 epochs using a batch size of 512. Early stopping was employed to prevent over-fitting if the validation loss stops improving for a duration of 5 or more epochs.\\

\subsection{The Proposed Ensemble Model}

\noindent Ensembling \cite{Zhang2012} is a widely-used technique known to improve a decision system's robustness and accuracy. The motivation for using ensemble models is to reduce the generalization error of predictions, as long as the base models are diverse and independent \cite{Kotu2019}. Algorithms such as Gradient Boosting and Random Forest utilize ensemble methods to combine the individual results of a large number of independent decision trees into a single prediction. This approach has been further extended to combine the predictions of unique, individual machine learning models via majority voting, weighted-scoring and other blending techniques to deliver more powerful and robust models.\\

\noindent To apply ensembling techniques to our own work, we blended the predictions from our XGB model with those from our ANN model using simple averaging. This resulted in four new experiments as follow:

\begin{enumerate}
\setcounter{enumi}{4}
  \item Train the proposed ensemble model (XGB + ANN) on SWELL and validate on NEURO and WESAD with no feature-engineering, using EDA and HR biomarkers only.
  \item Train the proposed ensemble model (XGB + ANN) on SWELL and validate on NEURO and WESAD with feature-engineering (10 total features), as explained in subsection \ref{sec:modelgen}.
  \item Train the proposed ensemble model (XGB + ANN) on StressData using LOSO validation and feature-engineering (10 total features), as explained in subsection \ref{sec:modelgen}.
  \item Train the proposed ensemble model (XGB + ANN) on StressData (excluding WESAD) and validate on WESAD with feature-engineering (10 total features), as explained in subsection \ref{sec:modelgen}.
\end{enumerate}

\subsection{Class Balancing} \label{sec:classbalancing}
\noindent Reviewing the initial results from experiment 7 when using XGB, ANN and their ensemble against the labeled metric for all 99 StressData subjects, we noted predictions for UBFC-Phys vastly outperformed those for WESAD, SWELL and NEURO across both XGB and ANN models. This was attributed to two potential factors; i) the substantially larger subject size of the UBFC-Phys dataset and ii) the substantially shorter recording segment size of UBFC-Phys samples.\\

\noindent We further noted a reduction in predictive accuracy in Experiment 8, when the WESAD dataset was excluded from StressData and used as a new, unseen validation dataset. This implies that LOSO validation alone may not be a good predictor of overall model generalization ability, likely due to an imbalance across both labeled metric and the significant variance in recording segment sizes across the datasets when merged into StressData.\\

\noindent To address this imbalance, we combined the EXAM dataset with StressData, resulting in a total of 129 subjects. Next, we split the stressed and non-stressed periodic segments of all 129 subjects into two groups. We noted the shortest segment across all subjects were from the UBFC-Phys dataset, at 201 seconds. We therefore further split each stressed and non-stressed subset across all subjects into smaller sets of 180 seconds each (3 minute intervals).\\

\noindent This resulted in a total of 3,758 samples of 3-minute segments, labeled as either stressed (2,962 samples) or non-stressed (796 samples). From these segments, random sampling can be performed to build a substantially larger balanced training dataset of any number and combination of segments. We named this new dataset SynthesizedStressData.\\ 

\noindent In addition to the 8 aforementioned experiments, we performed two final experiments:

\begin{enumerate}
\setcounter{enumi}{8}
  \item Train the proposed ensemble model (XGB + ANN) on SynthesizedStressData and validate using LOSO. For this experiment, 200 training subjects were generated by using random sampling from the 3,758 segments to simulate two physiological states: a 6-minute long non-stressed baseline period, followed by a 6-minute long stressed period. Feature engineering was applied as explained in subsection \ref{sec:modelgen}.
  \item Train the proposed ensemble model (XGB + ANN) on SynthesizedStressData (excluding the WESAD data) and test on the WESAD dataset for validation. For this experiment, all WESAD samples were first excluded from SynthesizedStressData, resulting in a smaller total of 115 subjects. To emulate the WESAD experimental protocol \cite{Schmidt2018}, 200 training subjects were generated using random sampling to emulate two physiological states: a 12-minute long non-stressed period, followed by a 12-minute long stressed period. Feature engineering was applied as explained in subsection \ref{sec:modelgen}.\\
\end{enumerate}

\section{Results and Discussion}

\subsection{Generalization Study}

\noindent Experiments 1 to 2 tested two approaches, the first excluding feature engineering, with the second including feature engineering. These were performed for measuring generalization of models built on a single dataset (SWELL) containing a small number of test subjects, but with large amount of observations. These models were then validated against two new, unseen datasets (WESAD, NEURO) containing a substantially smaller amount of observations.\\

\noindent The results indicate low predictive power with the best model achieving 68\% accuracy using a 70/30 Train/Test split during training, with feature-engineering. More importantly, the results show that a substantially larger number of observations (from a small number of test subjects) within the training dataset did not assist in generalization.\\

\begin{figure}[!h]
\centering
\fbox{\includegraphics[width=\textwidth]{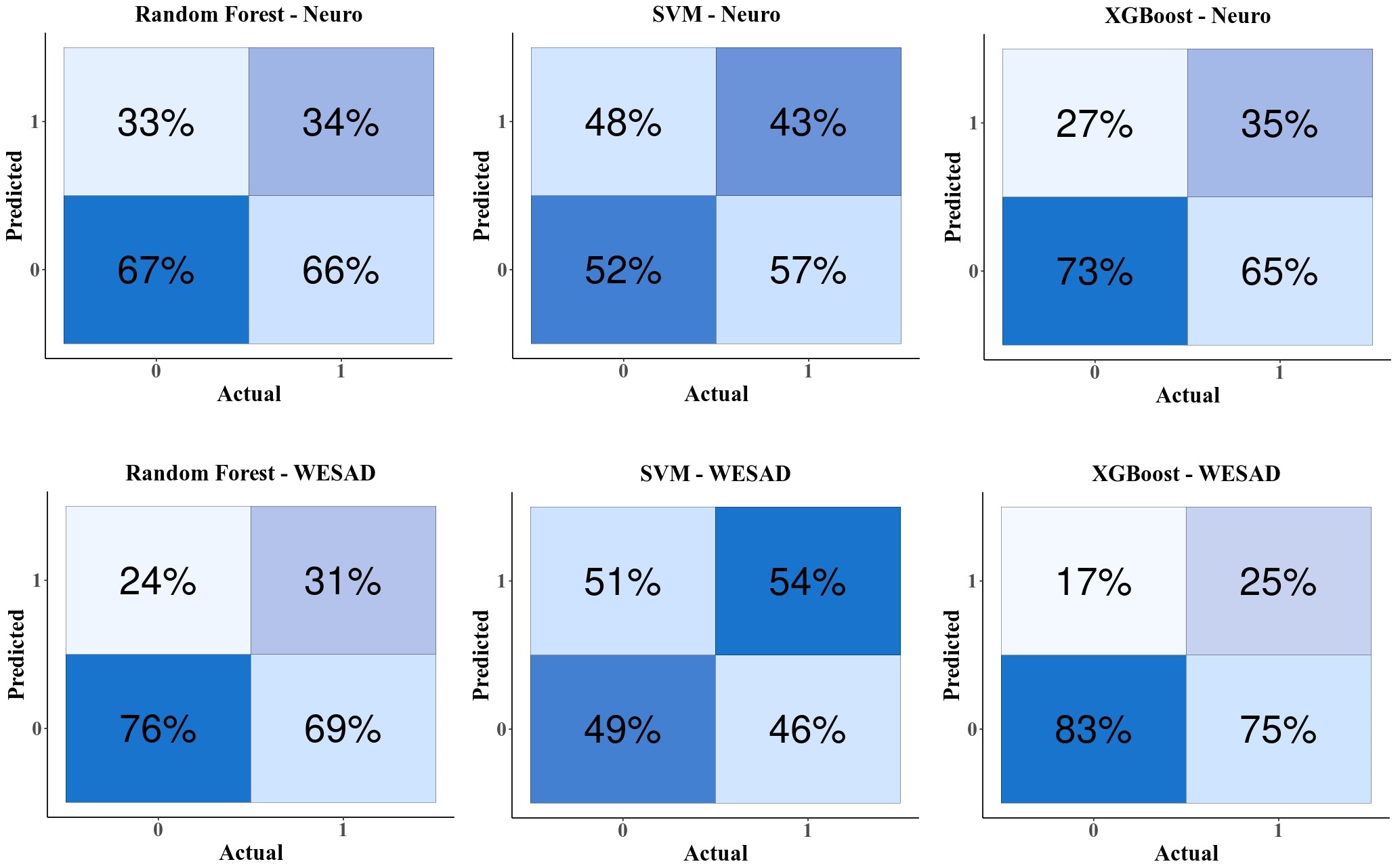}}
\caption{\label{fig:figure8}Confusion Matrix for Random Forest, SVM and XGBoost (Experiment 1).}%
\end{figure}

\noindent Furthermore, all three models (Random Forest, SVM, XGBoost) reported a large number of false negatives (Type II errors) and failed to predict the stress state of the test subjects (Figure \ref{fig:figure8}), with the SVM model performing slightly better compared to Random Forest and XGBoost.

\FloatBarrier

\begin{figure}[!h]
\centering
\fbox{\includegraphics[width=\textwidth]{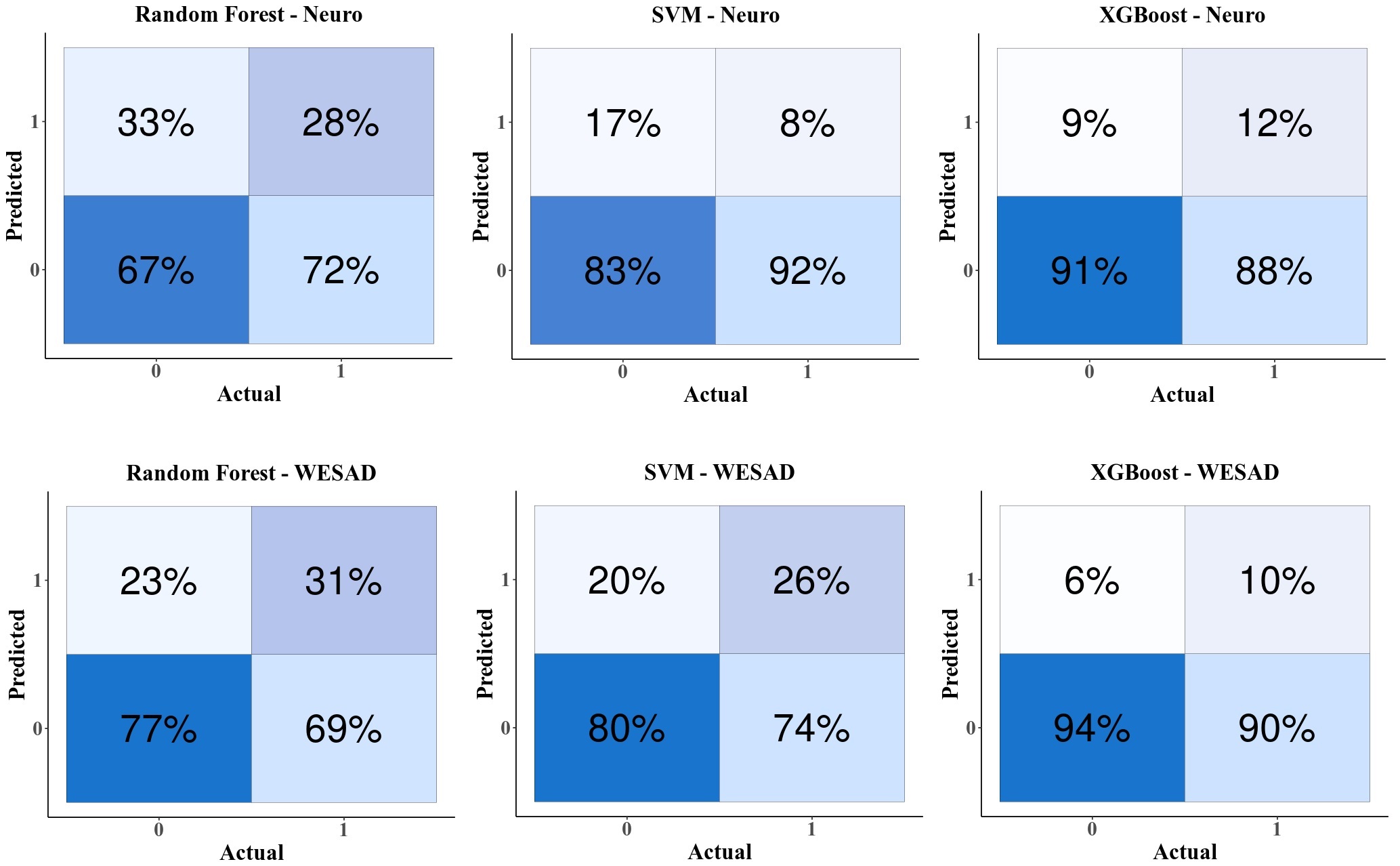}}
\caption{\label{fig:figure9}Confusion Matrix for Random Forest, SVM and XGBoost with engineered features (Experiment 2).}%
\end{figure}

\noindent Experiment 2 repeats Experiment 1 by utilizing feature-engineering. Accuracy, Precision, Recall and F1-scores improve slightly when predicting on the unseen WESAD dataset, but reduces for the NEURO dataset. Additionally, the resulting confusion matrix (Figure \ref{fig:figure9}) shows all three models producing a larger number of false negatives (Type II errors), compared to no feature-engineering as with Experiment 1.

\FloatBarrier

\begin{figure}[!h]
\centering
\fbox{\includegraphics[width=\textwidth]{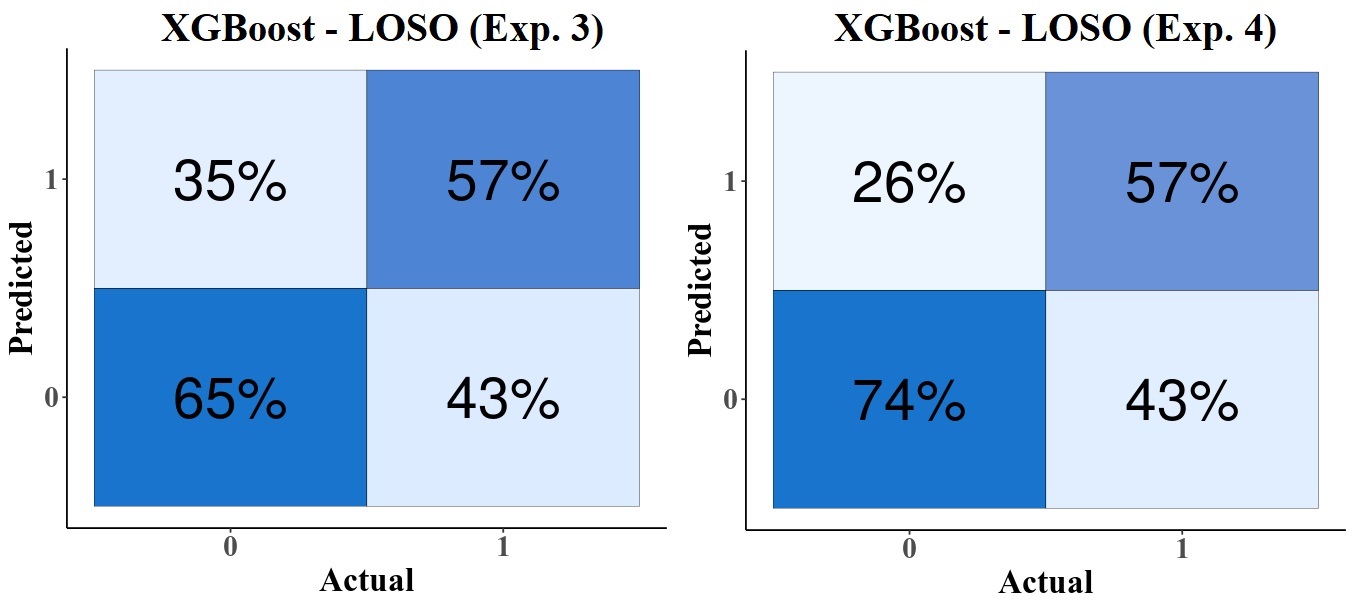}}
\caption{\label{fig:figure10}Confusion Matrix for Experiment 3 and 4.}%
\end{figure}

\noindent Once the datasets are merged to form the larger StressData dataset for training a machine learning model, predictive accuracy of 63\% is achieved with no feature engineering (Experiment 3), increasing to 72.36\% (Experiment 4) when using feature engineering, both using LOSO validation. Importantly, Type II errors have been substantially reduced with the introduction of a larger, merged dataset, with a slight improvement when using engineered features, as shown in Figure \ref{fig:figure10}. As neither Random Forest nor SVM showed a marked increase in performance over XGBoost, and due to the efficient training speed of XGBoost, both Random Forest and SVM was excluded for Experiments 5 to 10.\\

\FloatBarrier

\subsection{Ensemble Model}

\noindent Repeating Experiments 1 and 2 but using a weighted ensemble approach (XGB + ANN), predictive accuracy slightly improves to 51\% (NEURO) and 70\% (WESAD) for Experiment 5, with no feature engineering, then 58\% (NEURO) and 68\% (WESAD) for Experiment 6, with feature engineering applied, when training on only the SWELL dataset. Experiment 7 combines the benefits of a larger dataset (StressData) with feature engineering and a weighted ensemble approach, reporting predictive accuracy of 80.33\% when using LOSO validation.\\

\begin{figure}[!h]
\centering
\fbox{\includegraphics[width=\textwidth]{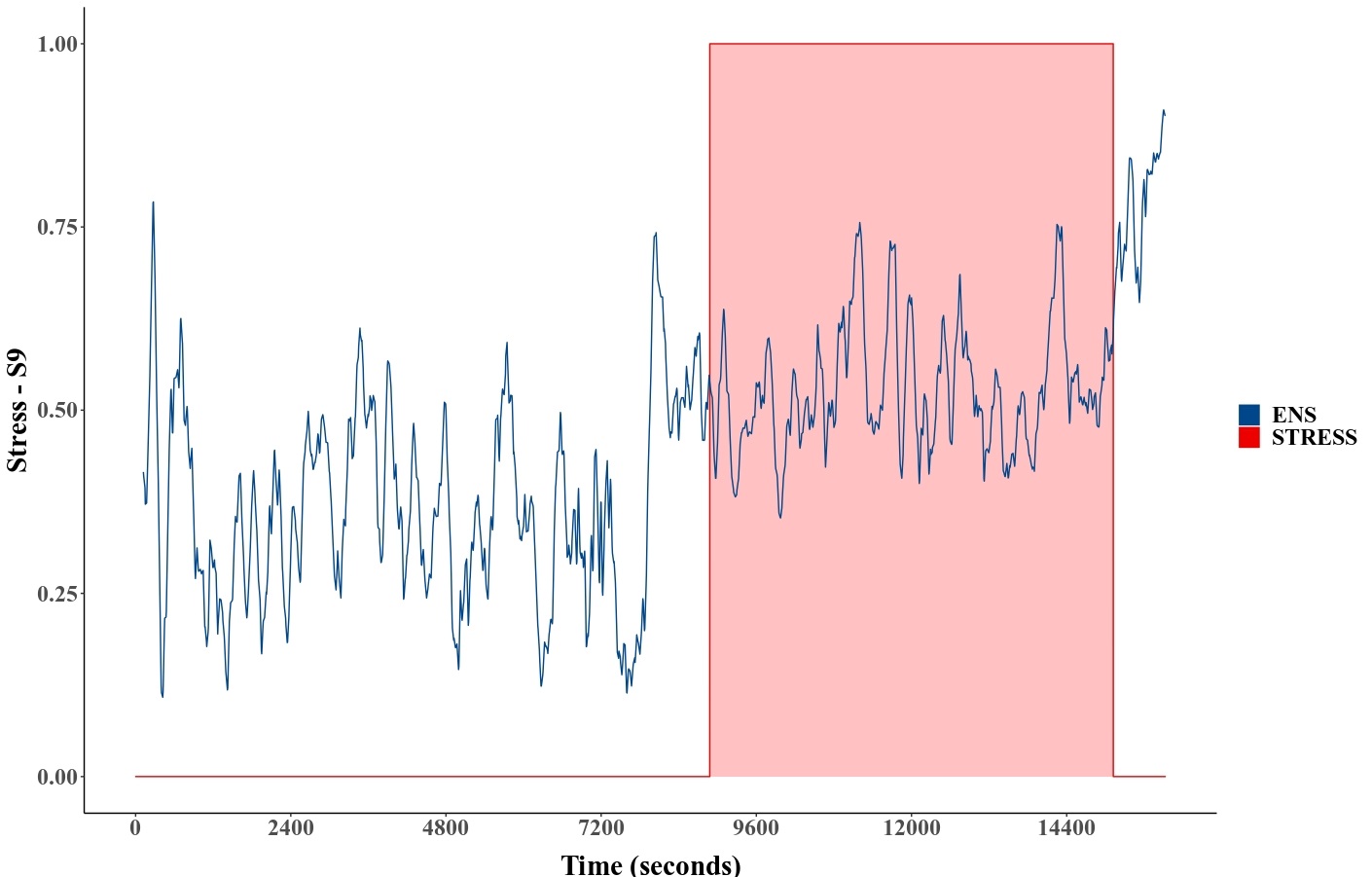}}
\caption{\label{fig:figure11} Ensemble model predictions (blue) on subject S9 from the SWELL dataset (Experiment 7).}%
\end{figure}

\noindent Figure \ref{fig:figure11} shows the numerical prediction plot for a randomly selected subject (S9) from the SWELL dataset (Experiment 7), with red detailing the stress period and blue indicating the ensemble predictions. The result shows good correlation with the period labeled as stressed. We note from 15,600 seconds onward, the subject is still showing an elevated level of stress, while the labeled recording period has been marked as non-stressed, likely not allowing for a cool down stage between stress and non-stress periods during biomarker recording.\\

\FloatBarrier

\begin{figure}[t]
\centering
\fbox{\includegraphics[width=\textwidth]{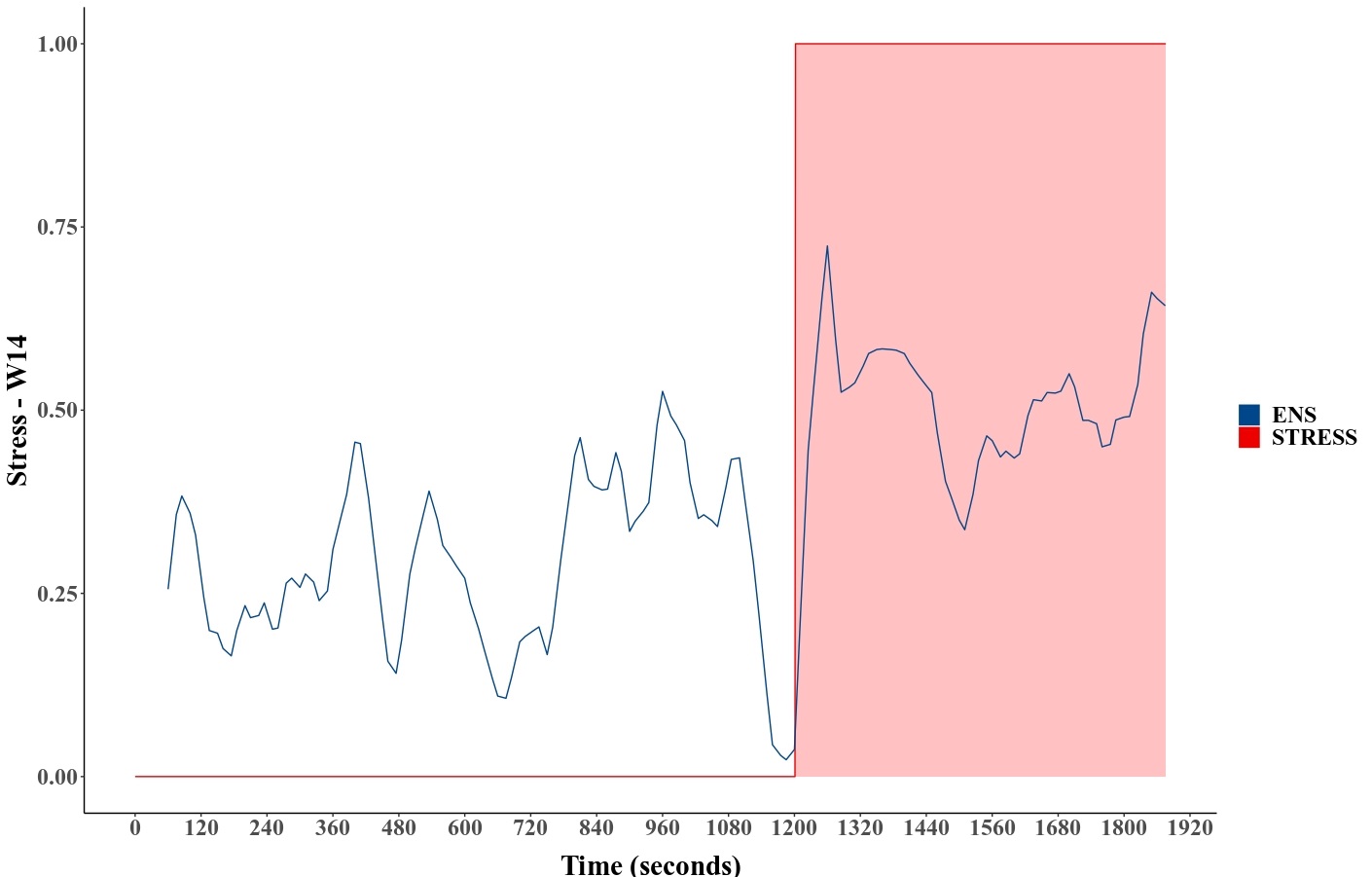}}
\caption{\label{fig:figure12}Ensemble model predictions (blue) on subject W14 from the WESAD dataset (Experiment 7).}%
\end{figure}

\noindent Figure \ref{fig:figure12} shows the numerical prediction plot for a randomly selected subject (here W14) from the WESAD dataset (Experiment 7), with red detailing the stress period and blue indicating the ensemble predictions. This result shows the ensemble model reacting well when subjects are subjected to a period of stress during the experiment.\\

\noindent Comparing the results to those from Experiments 1 to 4, we note slightly better results for the ensemble (using LOSO validation) when both a larger, more varied training dataset is utilized in combination with feature engineering. Various combinations of weighting were tested and utilized in order to achieve the highest accuracy rates, with the XGB model generally outperforming the ANN in most experiments:

\begin{itemize}
  \item Experiment 5 - 60\% (XGB) and 40\% (ANN).
  \item Experiment 6 - 80\% (XGB) and 20\% (ANN).
  \item Experiment 7 - 40\% (XGB) and 60\% (ANN) when predicting on biomarker segments exceeding 16 minutes in duration, and 70\% (XGB) and 30\% (ANN) for biomarker segments shorter than 16 minutes in duration.
  \item Experiment 8 - 30\% (XGB) and 70\% (ANN).
\end{itemize}

\noindent However, when training StressData with the WESAD dataset excluded (Experiment 8), predictive accuracy reduced to 59\% when testing on WESAD as an unseen validation set. Various class balancing methods were applied to StressData including over, under and both-sampling across the labeled stress metric, with both-sampling producing the highest score of 59\%, compared to over and under-sampling. This lower accuracy result implies that LOSO validation alone (Experiment 7) is not a good measure of model generalization, as the training process includes the full StressData dataset, apart from the single subject being left out per training and scoring iteration. This result further implies that training on disparate datasets merged without extensive class and feature balancing across both labeled metric as well as the recording segment size of biomarkers, may not result in a well-generalized model.\\ 

\subsection{Ensemble Model with Synthetic Data Generation}

\noindent Next, in Experiment 9, the EXAM dataset was included into StressData, constituting a total of 129 unique subjects, forming SynthesizedStressData. Random sampling was then utilized for class balancing as described in section \ref{sec:classbalancing} to construct 200 training subjects consisting of two physiological states: 6 minutes of a non-stressed baseline, and 6 minutes of a stressed period, sampled from the pool of 3,758 3-minute StressData segments (2962 stressed and 796 non-stressed). Validation was performed using LOSO, achieving a mean accuracy of 89\% on this instance of  SynthesizedStressData.\\

\begin{figure}[t]
\centering
\fbox{\includegraphics[width=\textwidth]{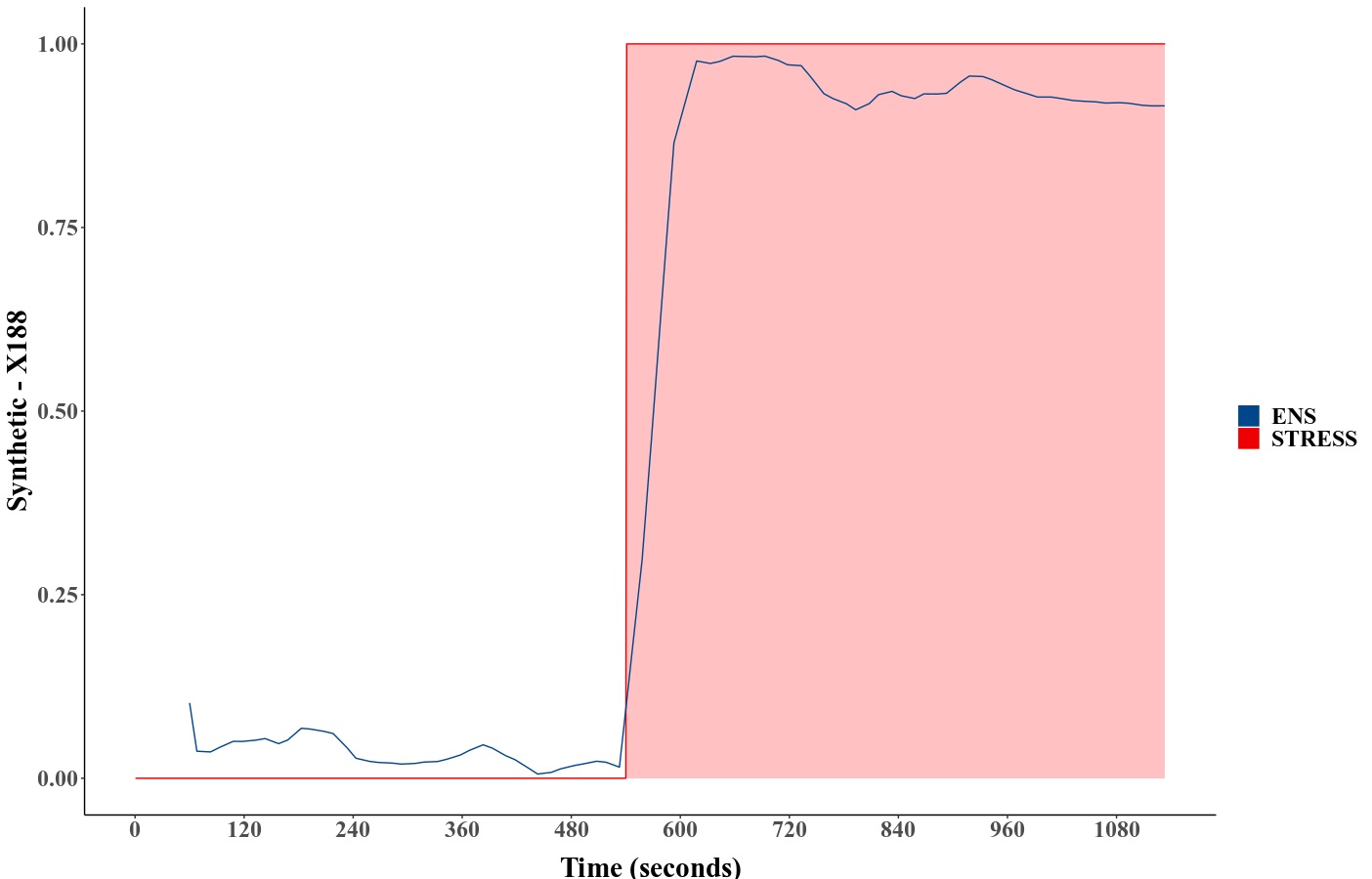}}
\caption{\label{fig:figure13}Predictions on subject X188 from Experiment 9.}%
\end{figure}

\noindent Precision, Recall and F1 score metrics are higher compared to experiments 1 through 8. Recall represents the model's ability to correctly predict the positives (stressed) out of actual positives (stressed), while Precision measures the number of predictions made that are actually positive (stressed), out of all positive predictions (stressed). In both cases, a higher score is desired. The F1 score is a good measure to use when seeking a balance between Precision and Recall. The prediction plot of subject X188 (from the SynthesizedStressData) is shown in Figure \ref{fig:figure13}, with a near-perfect match when predicting for both baseline and stressed periods.\\

\noindent Adjusting ensemble weighting based on varying biomarker segment sizes (Experiment 7) during the prediction phase, is not an ideal approach. Therefore, this method was not utilized in Experiment 9, where a fixed weighting of 60\% (XGB) and 40\% (ANN) (similar to Experiment 5) was  applied, given the strength of the XGB model over the ANN model in prior experiments.\\

\noindent Finally, for Experiment 10, the ensemble model trained on SynthesizedStressData was applied to the WESAD dataset to compare predictive power to those from Experiments 1, 2, 5, 6, and specifically Experiment 8. For this experiment, all WESAD data in StressData was removed and used as a validation set (similar to Experiment 8), resulting in a smaller total of 115 subjects used for training. Random sampling was again utilized to build a new instance of the SynthesizedStressData dataset as described in section \ref{sec:classbalancing}, to construct 200 training subjects consisting of two physiological states: 12 minutes of a non-stressed baseline, and 12 minutes of a stressed period, sampled from the pool of 3,600 3-minute segments (2917 stressed and 683 non-stressed).\\

\noindent A 12-minute interval was used to closely emulate the experimental protocol used when building the WESAD dataset, specifically the stressed period (approx. 10 minutes of stress \cite{Schmidt2018}). The random sampling, training and validation process was repeated for 10 iterations, with accuracy scores averaged across the 10 iterations. The results of these iterations are detailed in Table \ref{tab:resultsten}.\\

\noindent The ensemble method outperformed both the XGB and ANN models, with a mean accuracy rate of 85\%, and a very low standard deviation (0.029) across all 10 trials. A fixed weighting of 45\% (XGB) and 55\% (ANN) was used, showing more stability, and closer to an ideal balanced 50\%/50\% weighting. This result, compared to those from Experiment 8, shows a well generalized machine learning model capable of predicting accurately on new, unseen data (WESAD).\\

\begin{table}[t]
\centering
\caption{\label{tab:resultsten}Summary of accuracy scores for 10 trials of Experiment 10.}
\renewcommand{\arraystretch}{1.5}
\begin{tabular}{lcccccc}
\hline\hline
\textbf{Iteration} & \textbf{XGB}   & \textbf{ANN}   & \textbf{Ensemble} & \textbf{Precision} & \textbf{Recall} & \textbf{F1}     \\
1                  & 0.83           & 0.77           & 0.81              & 0.84               & 0.72            & 0.81            \\
2                  & 0.83           & 0.84           & 0.83              & 0.89               & 0.70            & 0.74            \\
3                  & 0.81           & 0.83           & 0.85              & 0.94               & 0.72            & 0.77            \\
4                  & 0.82           & 0.88           & 0.82              & 0.90               & 0.69            & 0.73            \\
5                  & 0.84           & 0.84           & 0.85              & 0.92               & 0.75            & 0.78            \\
6                  & 0.84           & 0.89           & 0.86              & 0.95               & 0.77            & 0.82            \\
7                  & 0.85           & 0.93           & 0.89              & 0.93               & 0.86            & 0.88            \\
8                  & 0.81           & 0.83           & 0.84              & 0.89               & 0.74            & 0.78            \\
9                  & 0.83           & 0.87           & 0.87              & 0.92               & 0.80            & 0.85            \\
10                 & 0.80           & 0.82           & 0.80              & 0.83               & 0.61            & 0.66            \\
\hline
\rowcolor[rgb]{0.753,0.753,0.753} \textbf{MEAN}      & \textbf{0.82}  & \textbf{0.82}  & \textbf{0.85}     & \textbf{0.90}      & \textbf{0.74}   & \textbf{0.78}   \\
\rowcolor[rgb]{0.753,0.753,0.753} \textbf{SD}       & \textbf{0.015} & \textbf{0.045} & \textbf{0.029}    & \textbf{0.041}     & \textbf{0.065}  & \textbf{0.062} \\
\hline\hline
\end{tabular}
\end{table}

\begin{figure}
\centering
\fbox{\includegraphics[width=\textwidth]{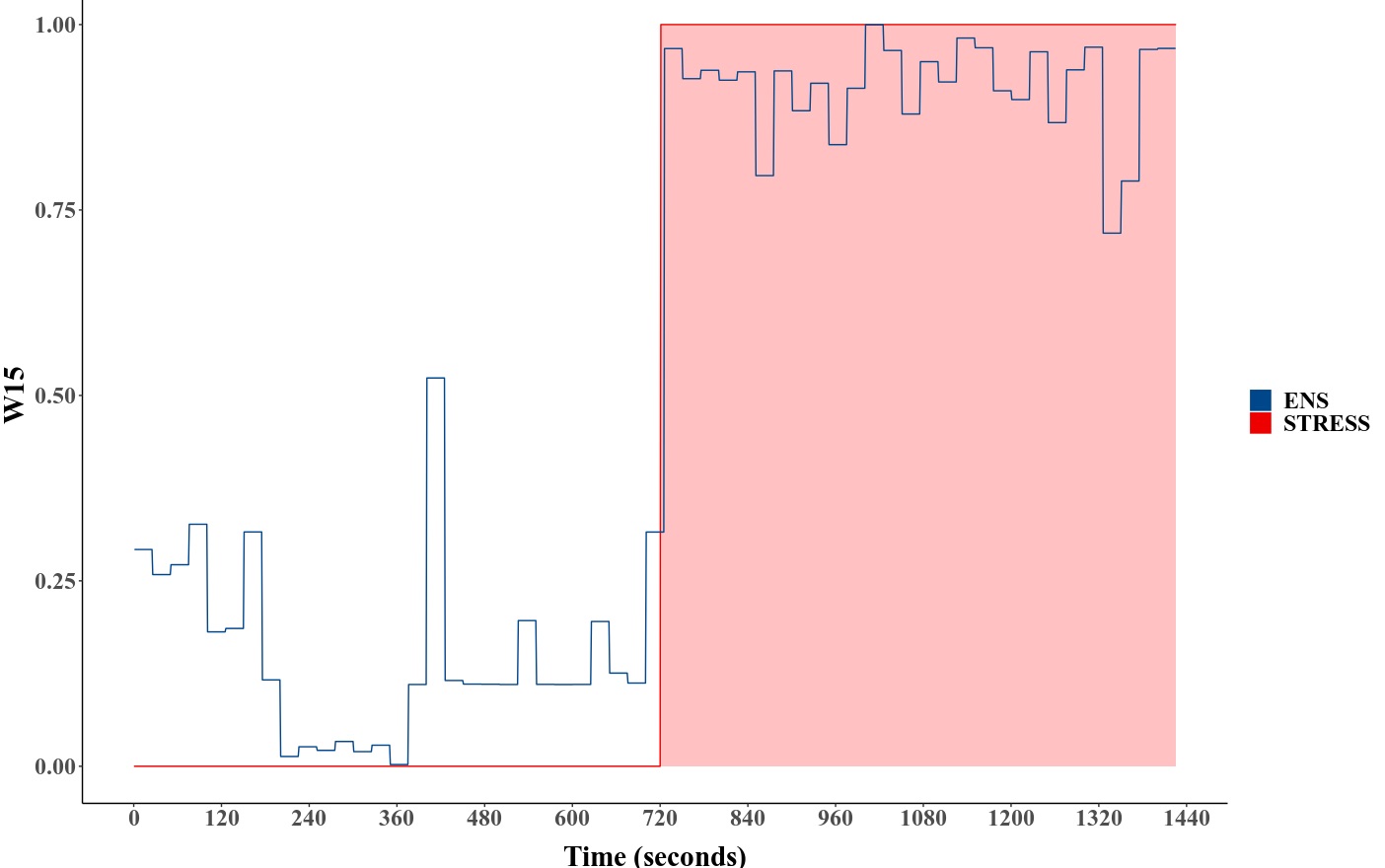}}
\caption{\label{fig:figure14}Predictions on subject W15 from the WESAD dataset (Experiment 10).}%
\end{figure}

\noindent Figure \ref{fig:figure14} shows a prediction plot of subject W15 from the WESAD dataset. We note a slight stress over-prediction during the baseline condition, with high correlation between prediction and label across the 24 minute experimental period.\\

\FloatBarrier

\begin{figure}
\centering
\fbox{\includegraphics[width=\textwidth]{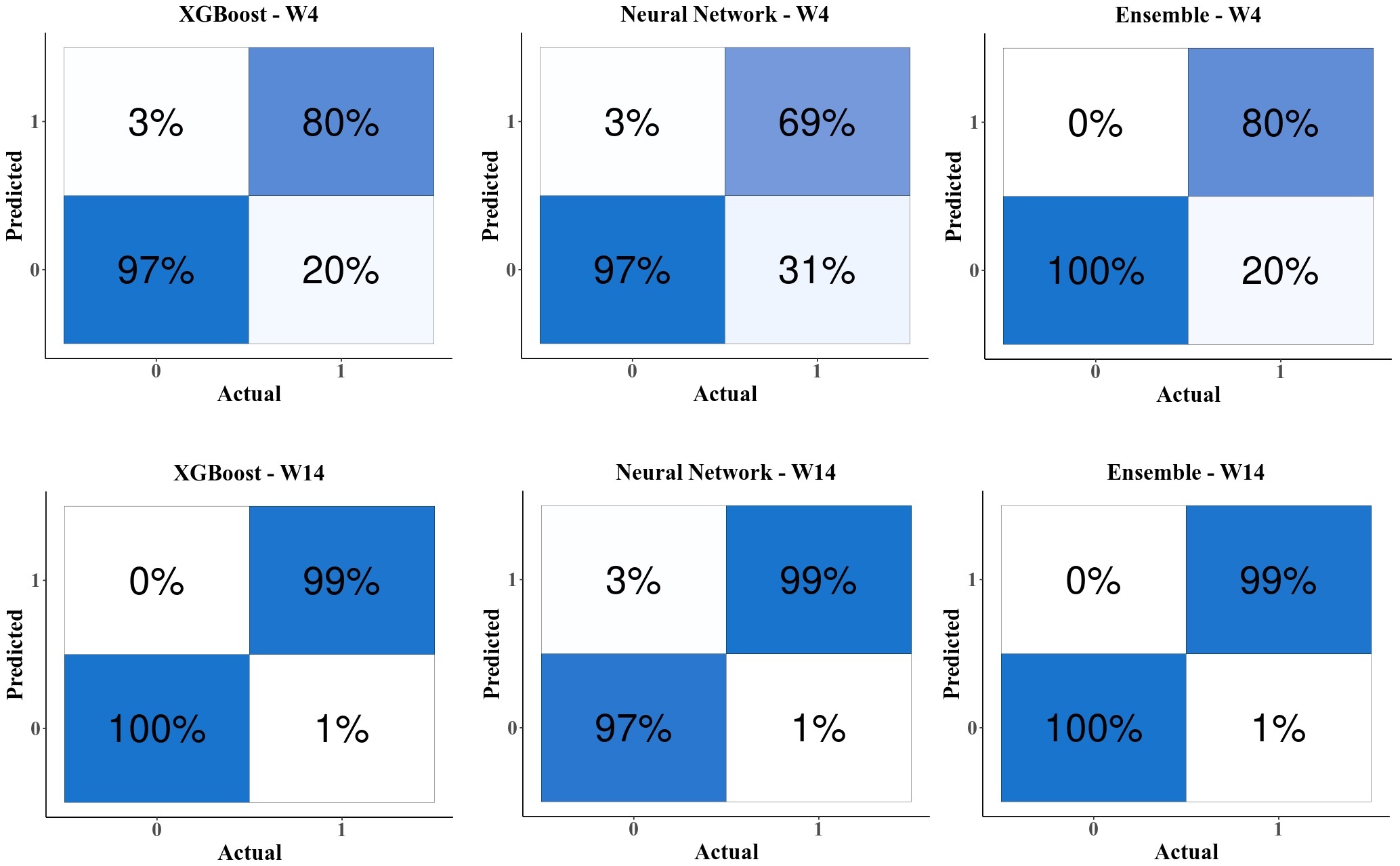}}
\caption{\label{fig:figure15}Confusion Matrix for subjects W4 and W14 (Experiment 10).}%
\end{figure}

\noindent Revisiting the findings from Experiment 1 through Experiment 4, we find a significant reduction in Type II errors as shown in the confusion matrix of all three models when plotted for test subjects W4 and W14 from the unseen WESAD dataset (Figure \ref{fig:figure15}). While both XGBoost and Ensemble models perform equally well for test subject W14, we note the Ensemble model outperforming both XGBoost and Artificial Neural Network models for test subject W4.\\

\FloatBarrier

\noindent To emulate a real-world scenario where a new, unseen dataset is provided with a potentially stressful experimental scenario, we trained and applied the full SynthesizedStressData (3758 samples, 129 subjects) ensemble model to the Toadstool dataset, using a balanced 50\% (XGB) and 50\% (ANN) weighting.\\
 
\begin{figure}[!h]
\centering
\fbox{\includegraphics[width=\textwidth]{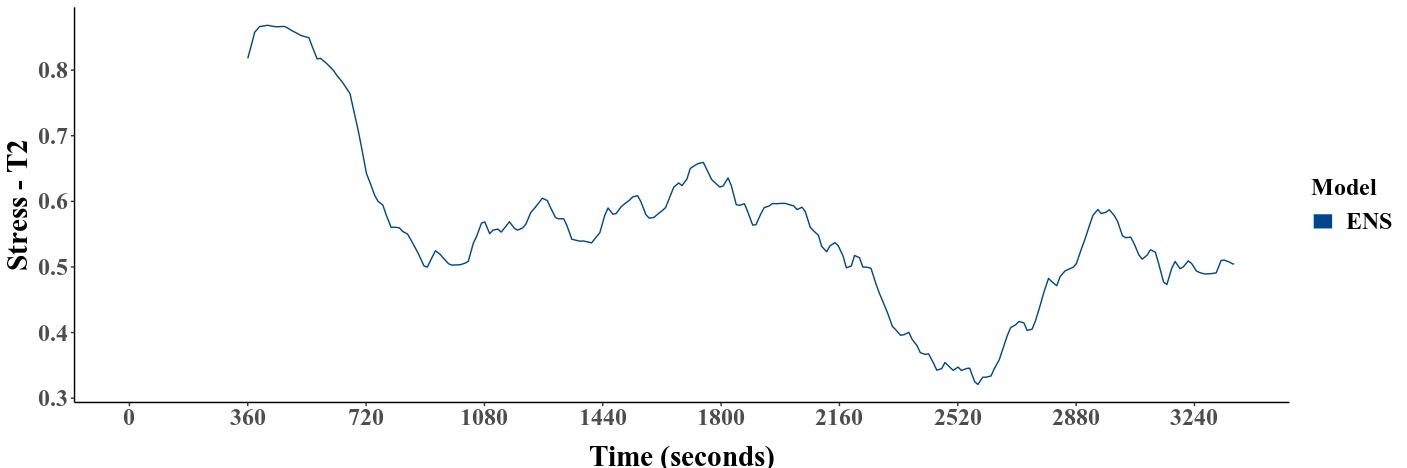}}
\caption{\label{fig:figure16}Predictions on subject T2 from the Toadstool dataset.}%
\end{figure}

\noindent This dataset contains sensor biomarker data recorded during periods of game-play where levels of subject stress may have been perceived as high. No labeling is provided for this dataset to compare predictive accuracy. Figure \ref{fig:figure16} shows a plot of the ensemble predictions for subject T2. We note elevated levels of stress when game-play starts, with a period of stabilization, prior to an increase as game-play reaches the end stage. While no definitive conclusions can be made from this test, it does show correlation between predictions and perceived levels of stress in this particular experimental scenario.

\FloatBarrier

\noindent The results from the ten aforementioned experiments are detailed in Table \ref{tab:results}. When examining the difference between experiment 4 and 7, we noted higher precision, indicating the ensemble model is better capable of predicting the stress state when subjects are in fact, under stress, while the improved recall indicates a better measure of identifying true positives (stressed condition). For experiments 2 and 6, there is a reduction in precision and recall for the WESAD dataset (better for the NEURO dataset), indicating that the ANN model, when part of the ensemble, fared worse for this dataset, reducing the overall precision and recall compared to singular models used for experiment 2. These differences when predicting on new, unseen data steered us towards merging the datasets and ultimately generating a synthetic dataset to address these shortcomings of training on small datasets.\\

\noindent  We also conducted an experiment where the large SWELL dataset was excluded from the SynthesizedStressData for training. As expected, this experiment yielded poor results due to the low number of subjects and observations leading to a lack of statistical significance. The result and the code for this supplementary experiment are available through the paper’s public code repository (see Supplement10.R).

\begin{sidewaystable}
\centering
\caption{\label{tab:results}Summary of experimental results.}
\renewcommand{\arraystretch}{1.5}
\resizebox{\textwidth}{!}{
\begin{tabular}{clllccccc}
\hline\hline
\textbf{Experiment}                  & \textbf{Model} & \textbf{Train Data} & \textbf{Validation Data} &  \textbf{F/E} & \textbf{Accuracy} & \textbf{Precision} & \textbf{Recall} & \textbf{F1}  \\
\hline
\rowcolor[rgb]{0.753,0.753,0.753} 1                              & XGB      & SWELL                 & NEURO, WESAD    &                            & 50\%, 66\%                   & 0.59, 0.27                    & 0.36, 0.29                 & 0.45, 0.28              \\
\rowcolor[rgb]{0.753,0.753,0.753} 1                              & SVM      & SWELL                 & NEURO, WESAD    &                            & 47\%, 50\%                   & 0.52, 0.31                    & 0.43, 0.54                & 0.47, 0.40              \\
\rowcolor[rgb]{0.753,0.753,0.753} 1                              & Random Forest      & SWELL                 & NEURO, WESAD    &                            & 49\%, 63\%                   & 0.55, 0.37                    & 0.34, 0.30                & 0.42, 0.33              \\
2                              & XGB      & SWELL                 & NEURO, WESAD    & $\bullet$ & 50\%, 68\%                   & 0.53, 0.36                    & 0.24, 0.59                 & 0.33, 0.45              \\
2                              & SVM      & SWELL                 & NEURO, WESAD    & $\bullet$ & 41\%, 63\%                   & 0.34, 0.35                    & 0.07, 0.25                 & 0.12, 0.30              \\
2                              & Random Forest      & SWELL                 & NEURO, WESAD    & $\bullet$ & 46\%, 62\%                   & 0.52, 0.33                    & 0.24, 0.22                 & 0.33, 0.26              \\
\rowcolor[rgb]{0.753,0.753,0.753} 3                              & XGB      & StressData            & LOSO            &                            & 63.00\%                      & 0.44                          & 0.68                       & 0.50                    \\
4                              & XGB      & StressData            & LOSO            & $\bullet$ & 72.36\%                      & 0.47                          & 0.66                       & 0.47                    \\
\rowcolor[rgb]{0.753,0.753,0.753} 5                              & Ensemble & SWELL            & NEURO, WESAD            & & 51\%, 70\%                      & 0.67, 0.55                          & 0.30, 0.44                       & 0.38, 0.56                    \\
6                              & Ensemble & SWELL            & NEURO, WESAD            & $\bullet$ & 50\%, 69\%                      & 0.81, 0.31                          & 0.34, 0.56                       & 0.43, 0.34                    \\
\rowcolor[rgb]{0.753,0.753,0.753} 7                              & Ensemble & StressData            & LOSO            & $\bullet$ & 80.33\%                      & 0.56                          & 0.75                       & 0.47                    \\
8                              & Ensemble & StressData (excl. WESAD)  & WESAD            & $\bullet$ & 59.00\%                      & 0.33                          & 0.22                       & 0.24                    \\

\rowcolor[rgb]{0.753,0.753,0.753} 9                              & Ensemble & SynthesizedStressData  & LOSO            & $\bullet$ & 89.00\%                      & 0.90                          & 0.88                       & 0.89                    \\
10                              & Ensemble & SynthesizedStressData (excl. WESAD) & WESAD           &  $\bullet$ & 85.00\%                         & 0.90                          & 0.74                       & 0.78                          \\

\hline\hline
\end{tabular}
}
\end{sidewaystable}
\FloatBarrier

\section{Conclusions and Discussion}

\noindent In this work, we evaluated the generalization ability of machine learning models trained on datasets containing stress biomarker data for a small number of study subjects. We found these models performed poorly, and proposed a methodology to engineer merging several small datasets into a single larger dataset which, when combined with feature-engineering, delivered a more robust stress detection model with significantly improved performance. This study was limited to utilizing the EDA and HR biomarkers generally available from medical-grade wearable devices such as the Empatica E4, for which a number of datasets are publicly available and specifically labeled for stress. While several commercially available devices are capable of measuring and recording additional sensor data including skin temperature, blood-volume-pulse and data from an accelerometer, these signals were excluded for this study as both accelerometer and temperature data are likely to be influenced by the study setting and experimental environment.\\

\noindent We found the recommendations by Brysbaert \emph{et al.} \cite{Brysbaert2019}, with regards to number of study subjects required to achieve sufficient statistical power (\textgreater 85\%) to correlate with our own findings, and showed model generalization to occur with more test subjects, compared to low generalization for a smaller number of subjects, as typically found in publicly available stress datasets. This finding has important implications for machine learning researchers who would normally expect generalization to occur given a sufficient number of observations, regardless of the number of individual test subjects that contributed to those observations. However, when the problem domain relates to behavioral science or a dependency on physiological biomarker data, a varied dataset consisting of biomarker data recorded from a sufficiently varied number of test subjects is equally important as, or more important than, dataset size and observation count alone.\\

\noindent While prior studies \cite{Nkurikiyeyezu2019} suggested that generalization is only possible with personalized models built for individual subjects, we found that we could reproduce our results on new, unseen data. Stress is not a binary condition, and the approach developed in this study delivers a methodology for predicting both a binary condition, as well as a level of stress experienced via regression, with accuracy levels \textgreater 85\%. A substantial challenge in using public datasets for training machine learning models to predict physiological responses, such as acute stress, are the varied experimental protocols used during recording, combined with large variance in recorded biomarker segment size. This can lead to class imbalance which is hard to overcome using traditional methods such as over and/or under-sampling of observations used for training.\\

\noindent Finally, we demonstrated the use of ensemble methods to combine the individual predictive power of unique machine learning algorithms, into a more powerful and robust stress detection model. By utilizing non-complex feature-engineering techniques with a tumbling window of 25 seconds, we presented a methodology that can deliver near real-time stress prediction from wearable sensor biomarker data. We further utilized random sampling of very small segments of either baseline or stressed periods, to build a balanced training dataset that performed well when using both LOSO validation, or when predicting on a new, unseen dataset (WESAD).\\

\noindent While the methods presented is not a substitute for large, validated and well-labeled datasets, the current lack of public data available for stress-related machine learning research limits the ability for researchers to reproduce prior findings \cite{Hullman2022}, and towards that end the full source code and feature-engineered StressData and SynthesizedStressData datasets used in this study are made available for future research use via the code-sharing platform GitHub at https://github.com/xalentis/Stress. We believe that the results and data presented herein can help advance the emerging field of machine learning for stress measurement from wearable devices. \\

\noindent The provided source code helps future studies implement the proposed random sampling methodology by  integrating their Empatica E4 dataset into our new synthesized dataset, or alternatively using the provided code as a guide when implementing the same approach for other devices and biomarkers. \\

\noindent The process of expanding the training dataset using our provided source code can lead to continual learning \cite{Ven2022}, which will further enhance dataset variability and help avoid future model drift. In the case of our study, techniques similar to domain-incremental learning can be utilized for incorporating additional datasets into the main SynthesizedStressData dataset, treating each as a different domain, before generating random synthetic samples for model re-training. Future studies could investigate the performance of popular continual learning techniques such as Elastic Weight Consolidation (EWC) and Gradient Episodic Memory (GEM) \cite{lopez2017gradient} against our proposed ensemble learning technique.  \\

 \bibliographystyle{elsarticle-num} 
 \bibliography{cas-refs}

\begin{thebibliography}{10}
\expandafter\ifx\csname url\endcsname\relax
  \def\url#1{\texttt{#1}}\fi
\expandafter\ifx\csname urlprefix\endcsname\relax\def\urlprefix{URL }\fi
\expandafter\ifx\csname href\endcsname\relax
  \def\href#1#2{#2} \def\path#1{#1}\fi

\bibitem{McEwen1998}
B.~S. McEwen, Protective and damaging effects of stress mediators, New England Journal of Medicine 338~(3) (1998) 171--179.
\newblock \href {https://doi.org/10.1056/nejm199801153380307} {\path{doi:10.1056/nejm199801153380307}}.

\bibitem{McEwen1993}
B.~S. McEwen, E.~Stellar, {Stress and the Individual: Mechanisms Leading to Disease}, Archives of Internal Medicine 153~(18) (1993) 2093--2101.

\bibitem{McEwen2006}
B.~S. McEwen, Physiology and neurobiology of stress and adaptation: Central role of the brain, Physiological Reviews 87~(3) (2007) 873--904.

\bibitem{Sriramprakash2017}
S.~Sriramprakash, V.~D. Prasanna, O.~R. Murthy, Stress detection in working people, Procedia Computer Science 115 (2017) 359--366.
\newblock \href {https://doi.org/10.1016/j.procs.2017.09.090} {\path{doi:10.1016/j.procs.2017.09.090}}.

\bibitem{Limas2018}
M.~A. Jiménez-Limas, C.~A. Ramírez-Fuentes, B.~Tovar-Corona, L.~I. Garay-Jiménez, Feature selection for stress level classification into a physiologycal signals set, in: 2018 15th International Conference on Electrical Engineering, Computing Science and Automatic Control (CCE), 2018, pp. 1--5.
\newblock \href {https://doi.org/10.1109/ICEEE.2018.8533968} {\path{doi:10.1109/ICEEE.2018.8533968}}.

\bibitem{Schmidt2018}
P.~Schmidt, A.~Reiss, R.~Duerichen, C.~Marberger, K.~Van~Laerhoven, Introducing wesad, a multimodal dataset for wearable stress and affect detection, in: Proceedings of the 20th ACM International Conference on Multimodal Interaction, ICMI '18, Association for Computing Machinery, New York, NY, USA, 2018, p. 400–408.

\bibitem{Nkurikiyeyezu2019}
K.~Nkurikiyeyezu, A.~Yokokubo, G.~Lopez, The effect of person-specific biometrics in improving generic stress predictive models (2019).
\newblock \href {https://doi.org/10.48550/ARXIV.1910.01770} {\path{doi:10.48550/ARXIV.1910.01770}}.

\bibitem{Eskandar2020}
S.~Eskandar, S.~Razavi, Using deep learning for assessment of workers' stress and overload, in: Proceedings of the 37th International Symposium on Automation and Robotics in Construction (ISARC), International Association for Automation and Robotics in Construction (IAARC), Kitakyushu, Japan, 2020, pp. 872--877.
\newblock \href {https://doi.org/10.22260/ISARC2020/0120} {\path{doi:10.22260/ISARC2020/0120}}.

\bibitem{Siirtola2020}
P.~Siirtola, J.~Röning, Comparison of regression and classification models for user-independent and personal stress detection, Sensors 20~(16) (2020) 4402.
\newblock \href {https://doi.org/10.3390/s20164402} {\path{doi:10.3390/s20164402}}.

\bibitem{Indikawati2020g}
F.~I. Indikawati, S.~Winiarti, Stress detection from multimodal wearable sensor data, {IOP} Conference Series: Materials Science and Engineering 771~(1) (2020) 012028.
\newblock \href {https://doi.org/10.1088/1757-899x/771/1/012028} {\path{doi:10.1088/1757-899x/771/1/012028}}.

\bibitem{Li2020}
R.~Li, Z.~Liu, Stress detection using deep neural networks, {BMC} Medical Informatics and Decision Making 20~(S11) (dec 2020).
\newblock \href {https://doi.org/10.1186/s12911-020-01299-4} {\path{doi:10.1186/s12911-020-01299-4}}.

\bibitem{Iqbal2021g}
T.~Iqbal, P.~Redon-Lurbe, A.~J. Simpkin, A.~Elahi, S.~Ganly, W.~Wijns, A.~Shahzad, A sensitivity analysis of biophysiological responses of stress for wearable sensors in connected health, {IEEE} Access 9 (2021) 93567--93579.
\newblock \href {https://doi.org/10.1109/access.2021.3082423} {\path{doi:10.1109/access.2021.3082423}}.

\bibitem{Liapis2021}
A.~Liapis, E.~Faliagka, C.~P. Antonopoulos, G.~Keramidas, N.~Voros, Advancing stress detection methodology with deep learning techniques targeting {UX} evaluation in {AAL} scenarios: Applying embeddings for categorical variables, Electronics 10~(13) (2021) 1550.
\newblock \href {https://doi.org/10.3390/electronics10131550} {\path{doi:10.3390/electronics10131550}}.

\bibitem{Ninh2021}
V.-T. Ninh, S.~Smyth, M.-T. Tran, C.~Gurrin, Analysing the performance of stress detection models on consumer-grade wearable devices, in: Frontiers in Artificial Intelligence and Applications, {IOS} Press, 2021.
\newblock \href {https://doi.org/10.3233/faia210050} {\path{doi:10.3233/faia210050}}.

\bibitem{Khan2022}
N.~Khan, Semi-supervised generative adversarial network for stress detection using partially labeled physiological data (2022).
\newblock \href {https://doi.org/10.48550/ARXIV.2206.14976} {\path{doi:10.48550/ARXIV.2206.14976}}.

\bibitem{Empatica2022}
Empatica, \href{http://www.empatica.com}{Empatica | medical devices, ai and algorithms for remote patient monitoring} (1 2022).
\newline\urlprefix\url{http://www.empatica.com}

\bibitem{Kraaij2015}
W.~Kraaij, S.~Koldijk, M.~Sappelli, The swell knowledge work dataset for stress and user modeling research (2015).
\newblock \href {https://doi.org/10.17026/DANS-X55-69ZP} {\path{doi:10.17026/DANS-X55-69ZP}}.

\bibitem{Birjandtalab2016}
J.~Birjandtalab, D.~Cogan, M.~B. Pouyan, M.~Nourani, A non-eeg biosignals dataset for assessment and visualization of neurological status, in: 2016 IEEE International Workshop on Signal Processing Systems (SiPS), 2016, pp. 110--114.
\newblock \href {https://doi.org/10.1109/SiPS.2016.27} {\path{doi:10.1109/SiPS.2016.27}}.

\bibitem{Haouij2018}
N.~E. Haouij, J.-M. Poggi, S.~Sevestre-Ghalila, R.~Ghozi, M.~Jaïdane, {AffectiveROAD} system and database to assess driver{\textquotesingle}s attention, in: Proceedings of the 33rd Annual {ACM} Symposium on Applied Computing, {ACM}, 2018.
\newblock \href {https://doi.org/10.1145/3167132.3167395} {\path{doi:10.1145/3167132.3167395}}.

\bibitem{Svoren2020}
H.~Svoren, V.~Thambawita, P.~Halvorsen, P.~Jakobsen, E.~G. Ceja, F.~M. Noori, H.~L. Hammer, M.~Lux, M.~Riegler, S.~Hicks, Toadstool: A dataset for training emotional intelligent machines playing super mario bros (feb 2020).
\newblock \href {https://doi.org/10.31219/osf.io/4v9mp} {\path{doi:10.31219/osf.io/4v9mp}}.

\bibitem{Sabour2021}
R.~Meziati~Sabour, Y.~Benezeth, P.~De~Oliveira, J.~Chappe, F.~Yang, Ubfc-phys: A multimodal database for psychophysiological studies of social stress, IEEE Transactions on Affective Computing (2021) 1--1\href {https://doi.org/10.1109/TAFFC.2021.3056960} {\path{doi:10.1109/TAFFC.2021.3056960}}.

\bibitem{Amin2022}
M.~R. Amin, D.~Wickramasuriya, R.~T. Faghih, A wearable exam stress dataset for predicting cognitive performance in real-world settings (2022).
\newblock \href {https://doi.org/10.13026/KVKB-AJ90} {\path{doi:10.13026/KVKB-AJ90}}.

\bibitem{Hosseini2022}
S.~Hosseini, R.~Gottumukkala, S.~Katragadda, R.~T. Bhupatiraju, Z.~Ashkar, C.~W. Borst, K.~Cochran, A multimodal sensor dataset for continuous stress detection of nurses in a hospital, Scientific Data 9~(1) (jun 2022).
\newblock \href {https://doi.org/10.1038/s41597-022-01361-y} {\path{doi:10.1038/s41597-022-01361-y}}.

\bibitem{Brysbaert2019}
M.~Brysbaert, How many participants do we have to include in properly powered experiments? a tutorial of power analysis with reference tables, Journal of Cognition 2~(1) (2019).
\newblock \href {https://doi.org/10.5334/joc.72} {\path{doi:10.5334/joc.72}}.

\bibitem{mishra2020}
V.~Mishra, S.~Sen, G.~Chen, T.~Hao, J.~Rogers, C.-H. Chen, D.~Kotz, Evaluating the reproducibility of physiological stress detection models, Proc. ACM Interact. Mob. Wearable Ubiquitous Technol. 4~(4) (dec 2020).

\bibitem{BioSPPy}
I.~de~Telecomunicacoes, \href{https://biosppy.readthedocs.io/en/stable/}{Biosppy} (1 2015).
\newline\urlprefix\url{https://biosppy.readthedocs.io/en/stable/}

\bibitem{RSoftware}
{R Core Team}, \href{https://www.R-project.org/}{R: A Language and Environment for Statistical Computing}, R Foundation for Statistical Computing, Vienna, Austria (2021).
\newline\urlprefix\url{https://www.R-project.org/}

\bibitem{EmpaticaR}
A.~Henelius, \href{https://github.com/bwrc/empatica-r}{Empatica-r} (1 2015).
\newline\urlprefix\url{https://github.com/bwrc/empatica-r}

\bibitem{mishra2018}
V.~Mishra, G.~Pope, S.~Lord, S.~Lewia, B.~Lowens, K.~Caine, S.~Sen, R.~Halter, D.~Kotz, The case for a commodity hardware solution for stress detection, in: Proceedings of the 2018 ACM International Joint Conference and 2018 International Symposium on Pervasive and Ubiquitous Computing and Wearable Computers, UbiComp '18, Association for Computing Machinery, New York, NY, USA, 2018, p. 1717–1728.

\bibitem{LightGBM}
Microsoft, \href{https://lightgbm.readthedocs.io/en/latest/}{Lightgbm} (1 2022).
\newline\urlprefix\url{https://lightgbm.readthedocs.io/en/latest/}

\bibitem{CatBoost}
Yandex, \href{https://catboost.ai/}{Catboost} (1 2022).
\newline\urlprefix\url{https://catboost.ai/}

\bibitem{XGBoost}
T.~Chen, C.~Guestrin, \href{https://doi.org/10.1145/2939672.2939785}{Xgboost: A scalable tree boosting system}, KDD '16, Association for Computing Machinery, New York, NY, USA, 2016, p. 785–794.
\newblock \href {https://doi.org/10.1145/2939672.2939785} {\path{doi:10.1145/2939672.2939785}}.
\newline\urlprefix\url{https://doi.org/10.1145/2939672.2939785}

\bibitem{Latinne2001}
P.~Latinne, O.~Debeir, C.~Decaestecker, Limiting the number of trees in random forests, in: Multiple Classifier Systems, Springer Berlin Heidelberg, 2001, pp. 178--187.

\bibitem{Stathakis2009}
D.~Stathakis, How many hidden layers and nodes?, International Journal of Remote Sensing 30~(8) (2009) 2133--2147.
\newblock \href {https://doi.org/10.1080/01431160802549278} {\path{doi:10.1080/01431160802549278}}.

\bibitem{RELU}
K.~Fukushima, Visual feature extraction by a multilayered network of analog threshold elements, IEEE Transactions on Systems Science and Cybernetics 5~(4) (1969) 322--333.
\newblock \href {https://doi.org/10.1109/TSSC.1969.300225} {\path{doi:10.1109/TSSC.1969.300225}}.

\bibitem{Zhang2012}
C.~Zhang, Y.~Ma, Ensemble machine learning: Methods and applications, 2012.
\newblock \href {https://doi.org/10.1007/9781441993267} {\path{doi:10.1007/9781441993267}}.

\bibitem{Kotu2019}
V.~Kotu, B.~Deshpande, Data Science, Elsevier, 2019.
\newblock \href {https://doi.org/10.1016/c2017-0-02113-4} {\path{doi:10.1016/c2017-0-02113-4}}.

\bibitem{Hullman2022}
J.~Hullman, S.~Kapoor, P.~Nanayakkara, A.~Gelman, A.~Narayanan, The worst of both worlds: A comparative analysis of errors in learning from data in psychology and machine learning (03 2022).

\bibitem{Ven2022}
G.~M. van~de Ven, T.~Tuytelaars, A.~S. Tolias, Three types of incremental learning, Nature Machine Intelligence 4~(12) (2022) 1185--1197.
\newblock \href {https://doi.org/10.1038/s42256-022-00568-3} {\path{doi:10.1038/s42256-022-00568-3}}.

\bibitem{lopez2017gradient}
D.~Lopez-Paz, M.~Ranzato, Gradient episodic memory for continual learning, Advances in neural information processing systems 30 (2017).

\end{thebibliography}





\end{document}